\documentclass[10pt,twocolumn,letterpaper]{article}

\usepackage[pagenumbers]{cvpr} %

\usepackage{graphicx}
\usepackage{amsmath}
\usepackage{amssymb}
\usepackage{booktabs}

\usepackage[pagebackref,breaklinks,colorlinks]{hyperref}

\def\ie{\emph{i.e.}}
\def\eg{\emph{e.g.}}

\def\etal{{\em et al.}}

\usepackage{multirow}

\usepackage{enumitem}
\usepackage{tabularx}
\usepackage{multirow}
\usepackage{stfloats}
\usepackage{threeparttable}
\usepackage[usenames,dvipsnames]{xcolor}

\usepackage{bm}

\def\ie{\emph{i.e.}}
\def\eg{\emph{e.g.}}

\def\etal{{\em et al.}}

\newlength\savedwidth

\definecolor{darkergreen}{RGB}{21, 152, 56}
\definecolor{red2}{RGB}{252, 54, 65}
\newcommand\redp[1]{\textcolor{red2}{#1}}
\newcommand\greenp[1]{\textcolor{darkergreen}{#1}}

\usepackage{xcolor}
\definecolor{yl_color}{RGB}{128, 255, 0}

\definecolor{blue2}{RGB}{20, 54, 254}

\usepackage{bbding}

\usepackage[american]{babel}
\usepackage{microtype}
\usepackage{cleveref}

\usepackage{subcaption}

\usepackage[american]{babel}
\usepackage{microtype}

\usepackage{cleveref}
\crefname{section}{Sec.}{Secs.}
\Crefname{section}{Section}{Sections}
\Crefname{table}{Table}{Tables}
\crefname{table}{Tab.}{Tabs.}

\def\cvprPaperID{6688} %
\def\confName{CVPR}
\def\confYear{2023}

\begin{document}

\title{Turning a CLIP Model into a Scene Text Detector}

\author{
  Wenwen Yu\thanks{Equal contribution. ~~\textsuperscript{\dag}Corresponding author.} $  ^{,1}$,
  Yuliang Liu$^{*,1}$, 
  Wei Hua$^{1}$,
  Deqiang Jiang$^{2}$, 
  Bo Ren$^{2}$,
  Xiang Bai$^{\dagger,1}$      \\
    $^1$Huazhong University of Science and Technology ~~~ $^2$Tencent YouTu Lab\\
    \small{\texttt{\{wenwenyu,ylliu,whua\_hust,xbai\}@hust.edu.cn}}, \small{\texttt{\{dqiangjiang,timren\}@tencent.com}}
  }

\maketitle

\begin{abstract}
The recent large-scale Contrastive Language-Image Pretraining (CLIP) model has shown great potential in various downstream tasks via leveraging the pretrained vision and language knowledge. Scene text, which contains rich textual and visual information, has an inherent connection with a model like CLIP. 
Recently, pretraining approaches based on vision language models have made effective progresses in the field of text detection. 
In contrast to these works, this paper proposes a new method, termed TCM, focusing on Turning the CLIP Model directly for text detection without pretraining process. 
We demonstrate the advantages of the proposed TCM as follows: 
(1) The underlying principle of our framework can be applied to improve existing scene text detector.
(2) It facilitates the few-shot training capability of existing methods, \eg, by using 10\% of labeled data, we significantly improve the performance of the baseline method with an average of 22\% in terms of the F-measure on 4 benchmarks.
(3) By turning the CLIP model into existing scene text detection methods, we further achieve promising domain adaptation ability. The code will be publicly released at https://github.com/wenwenyu/TCM.

\end{abstract}

\section{Introduction}
\label{sec:intro}
Scene text detection
is a long-standing research topic aiming to localize the bounding box or polygon of each text instance from natural images, as it has wide practical applications scenarios, such as office automation, instant translation, automatic driving, and online education. With the rapid development of fully-supervised deep learning technologies, scene text detection has achieved remarkable progresses. 
Although supervised approaches have made remarkable progress in the field of text detection, they require extensive and elaborate annotations, \eg, character-level, word-level, and text-line level bounding boxes, especially polygonal boxes for arbitrarily-shaped scene text. Therefore, 
it is very important to investigate text detection methods under small amount of labeled data, \ie, few-shot training.

\begin{figure}[tbp]
\centering
\includegraphics[width=0.39\textwidth]{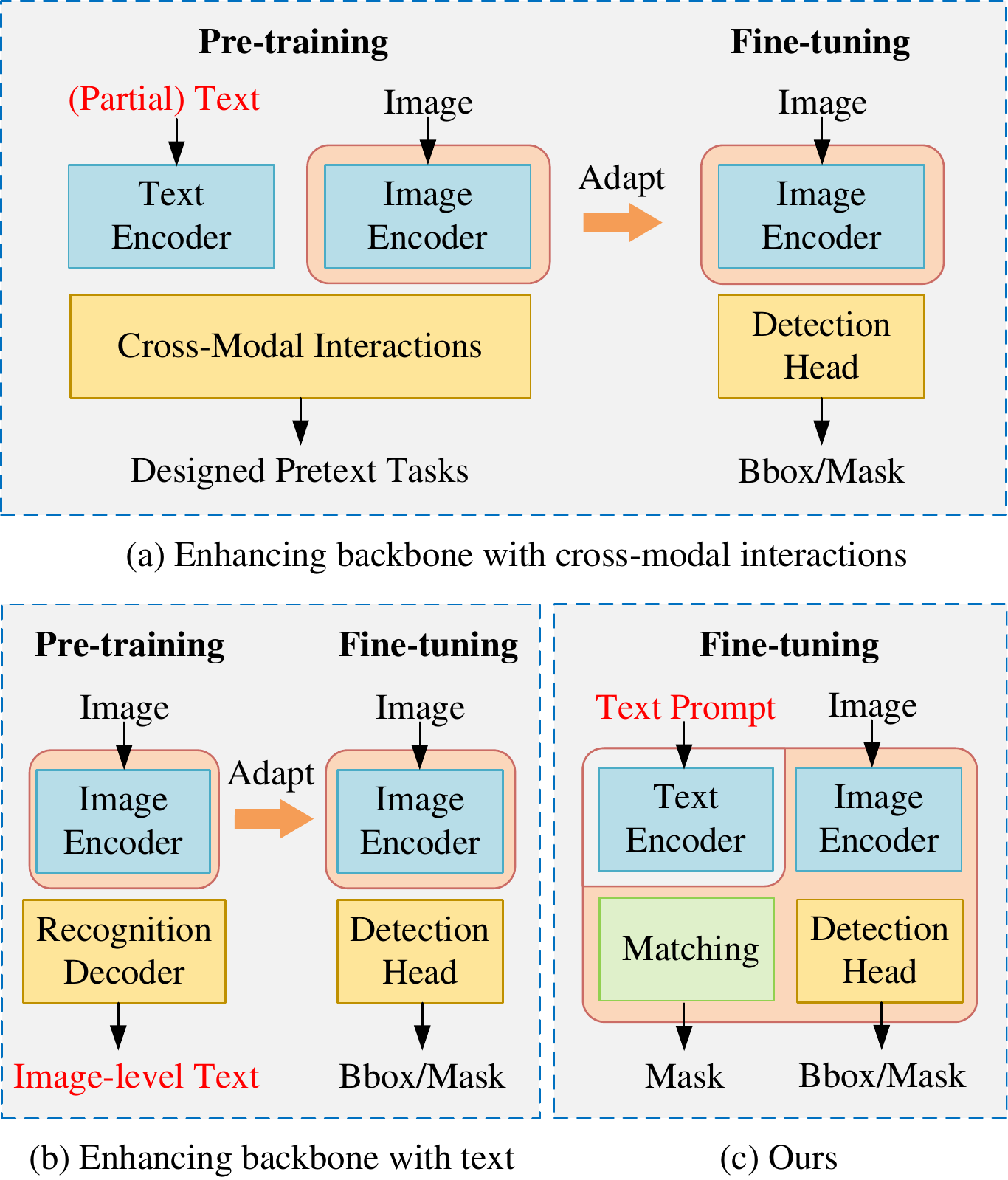}
\caption{Comparisons of different paradigms of using text knowledge for scene text detection. 
}
\vspace{-1em}
\label{fig:compare_intro}
\end{figure}

Recently, through leveraging the pretrained vision and language knowledge, the large-scale Contrastive Language-Image Pretraining (CLIP) model~\cite{Radford2021LearningTV} has demonstrated its significance in various downstream tasks. \eg, image classification~\cite{Zhou2022ConditionalPL}, object detection~\cite{Gu2022OpenvocabularyOD}, and semantic segmentation~\cite{Rao2022DenseCLIPLD, Xu2021ASB, Li2022LanguagedrivenSS}.

Compared to general object detection, scene text in natural images usually presents with both visual and rich character information, which has a natural connection with the CLIP model. Therefore, how to make full use of cross-modal information from visual, semantic, and text knowledge to improve the performance of the text detection models receives increasing attentions in recent studies. For examples, Song \etal~\cite{Song2022VisionLanguagePF}, inspired by CLIP, adopts fine-grained cross-modality interaction to align unimodal embeddings for learning better representations of backbone via carefully designed pretraining tasks.
Xue \etal~\cite{Xue2022LanguageMA} presents a weakly supervised pretraining method to jointly learn and align visual and partial textual information for learning effective visual text representations for scene text detection. Wan \etal~\cite{Wan2021SelfattentionBT} proposes self-attention based text knowledge mining to enhance backbone via an image-level text recognition pretraining tasks.

Different from these works, as shown in Figure~\ref{fig:compare_intro}, this paper focuses on turning the CLIP model for text detection without 
pretraining process. 
However, it is not trivial to incorporate the CLIP model into a scene text detector. The key is seeking a proper method to exploit the visual and semantic prior information conditioned on each image. 
In this paper, we develop a new method for scene text detection, termed as TCM, short for \textbf{T}urning a \textbf{C}LIP \textbf{M}odel into a scene text detector, which can be easily plugged to improve the scene text detection frameworks. 
We design a cross-modal interaction mechanism through visual prompt learning, which is implemented by cross-attention to recover the locality feature from the image encoder of CLIP to capture fine-grained information to respond to the coarse text region 
for the subsequent matching between text instance and language.
Besides, to steer the pretrained knowledge from the text encoder conditioned independently on different input images, we employ the predefined language prompt, learnable prompt, and a language prompt generator using simple linear layer to  get global image information.
In addition, we design an instance-language matching method to align the image embedding and text embedding, which encourages the image encoder to explicitly refine text regions 
from cross-modal visual-language priors. Compared to previous pretraining approaches, our method can be directly finetuned for the text detection task without 
pretraining process, 
as elaborated in Fig.~\ref{fig:compare_intro}. 
In this way, the text detector can absorb the rich visual or semantic information of text from CLIP. We summarize the advantages of our method as follows:

\begin{itemize}
\item We construct a new text detection framework, termed as TCM, 
which can be easily plugged to enhance the existing detectors.
\item Our framework can enable effective few-shot training capability. Such advantage is more obvious when using less training samples compared to the baseline detectors. Specifically, by using 10\% of labeled data, we improve the performance of the baseline detector by an average of 22\% in terms of the F-measure on 4 benchmarks.
\item TCM introduces promising domain adaptation ability, \ie, when using training data that is out-of-distribution of the testing data, the performance can be significantly improved. Such phenomenon is further demonstrated by a NightTime-ArT text dataset\footnote{\href{https://drive.google.com/file/d/1v3CshPqlvhpnK1_MKwqqkWJDikKl_g4Y}{NightTime-ArT Download Link}}, which we collected from the ArT dataset. 
\item Without pretraining process using specific pretext tasks, TCM can still leverage the prior knowledge from the CLIP model, outperforming previous scene text pretraining methods~\cite{Wan2021SelfattentionBT,Song2022VisionLanguagePF,Xue2022LanguageMA}.
\end{itemize}

\section{Related works}
\label{sec:rela}
\paragraph{Unimodal Scene Text Detection.} 
Unimodal scene text detection represents the method directly adopts the bounding boxes annotation only~\cite{Long2020SceneTD}. It can be roughly divided into two categories: Segmentation-based methods and regression-based methods. The segmentation-based methods usually conduct pixel-level~\cite{Liao2020RealtimeST,Tian2019LearningSE,Xue2019MSRMS,Li2019ShapeRT,Wang2019EfficientAA,Xie2019SceneTD,Liao2019RealTimeST}, segment-level~\cite{Shi2017DetectingOT,Long2018TextSnakeAF, Zhang2020DeepRR,Baek2019CharacterRA, Xu2019TextFieldLA,Tian2016DetectingTI,Tang2019SegLinkDD, Ye2020TextFuseNetST}, or  contour-level~\cite{Wang2020TextRayCG,Wang2020ContourNetTA} segmentation, then grouping segments into text instances via post-processing. 
The regression-based methods~\cite{zhu2021fourier,Zhang2016MultiorientedTD,He2017SingleST, He2017DeepDR, Liao2017TextBoxesAF, Zhou2017EASTAE, He2021MOSTAM,Zhang2019LookMT,Wang2019ArbitrarySS} regards text as a whole object and regress the bounding boxes of the text instances directly.

\paragraph{Cross-modal Assisted Scene Text Detection.} Unlike unimodal based scene text detection, cross-modal assisted scene text detection aims to make full use of cross-modal information including visual, semantic, and text knowledge to boost the performance. Wan \etal~\cite{Wan2021SelfattentionBT} utilized an image-level text recognition pretraining tasks to enhance backbone via the proposed self-attention based text knowledge mining mechanism.
Song \etal~\cite{Song2022VisionLanguagePF}, inspired by CLIP, designed three pretraining fine-grained cross-modality interaction tasks to align unimodal embeddings for learning better representations of backbone.
Xue \etal~\cite{Xue2022LanguageMA} jointly learned and aligned visual and partial text instances information for learning effective visual text representations via the proposed weakly supervised pretraining method. Long \etal~\cite{Long2022TowardsEU} proposed an end-to-end model to perform unified scene text detection and visual layout analysis simultaneously. The above methods explicitly leverage text or visual information to assist text detection. Instead, our method focuses on improving the performance results by turning a CLIP model into a scene text detector via leveraging pretrained text knowledge.

\section{Methodology}
\label{sec:method}
We begin by illustrating the CLIP model which we used for fetching the prior knowledge. Next, we introduce the technical details of TCM as well as the rationale behind it. An overview of our approach is shown in Fig.~\ref{fig:method_overall}.

\subsection{Contrastive Language-Image Pretraining} 
\begin{figure}[t]
\centering
\includegraphics[width=0.48\textwidth]{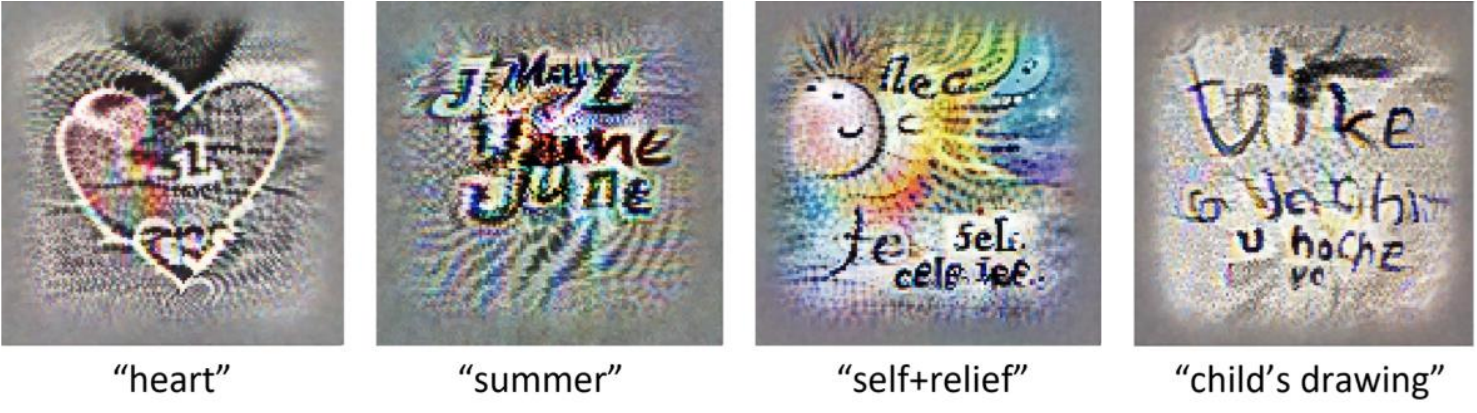}
\caption{The neurons in the clip model can directly respond to the text. The source images are from~\cite{goh2021multimodal}. }
\vspace{-1em}
\label{fig:clip_neuron}
\end{figure}

CLIP~\cite{Radford2021LearningTV}, which collects 400 million image-text pairs without human annotation for model pretraining, has well demonstrated the potential of learning transferable knowledge and open-set visual concepts. 
Previous study~\cite{goh2021multimodal} shows that different neurons in CLIP model can capture the corresponding concept literally, symbolically, and conceptually,
As shown in Fig.~\ref{fig:clip_neuron}, the CLIP model is an inborn text-friendly model which can effectively abstract the mapping space between image and text~\cite{Petroni2019LanguageMA}. 
During training, CLIP learns a joint embedding space for the two modalities via a contrastive loss. 
Given a batch of image-text pairs, for each image, CLIP maximizes the cosine similarity with the matched text while minimizing that with all other unmatched text. For each text, the loss is computed similarly as each image. In this way, CLIP can be used for zero-shot image recognition~\cite{Zhou2022ConditionalPL}. 
However, to exploit the relevant information from such a model, there are two prerequisites: 1) A proper method to effectively request the prior knowledge from the CLIP. 2) The original model can only measure the similarity between an integrated image and a single word or sentence. For scene text detection, there are usually many text instances per image, which are all required to be recalled equivalently.

\subsection{Turning a CLIP into a Text Detector}
\begin{figure}[htbp]
\centering
\includegraphics[width=0.46\textwidth]{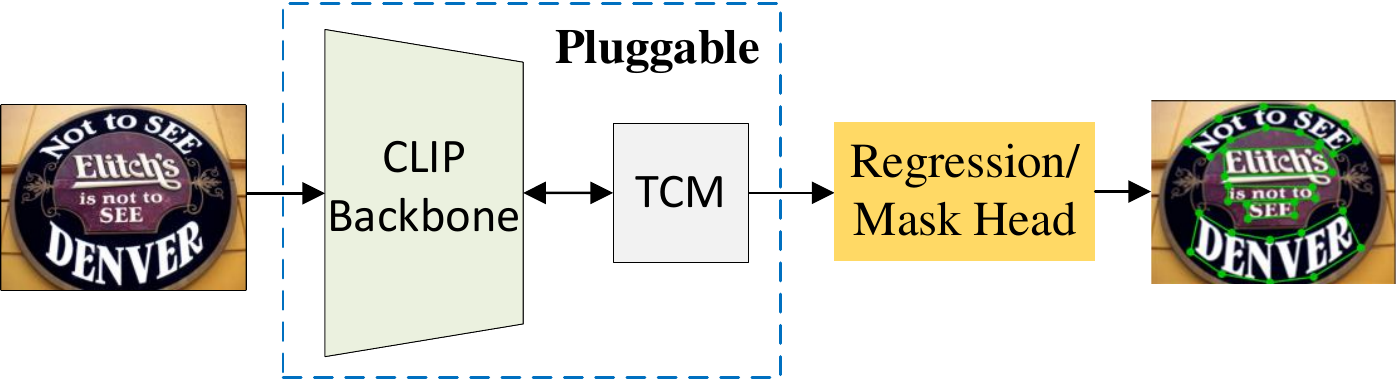}
\caption{The overall framework of our approach.}
\vspace{-2em}
\label{fig:method_overall}
\end{figure}

\begin{figure}[htbp]
\centering
\includegraphics[width=0.49\textwidth]{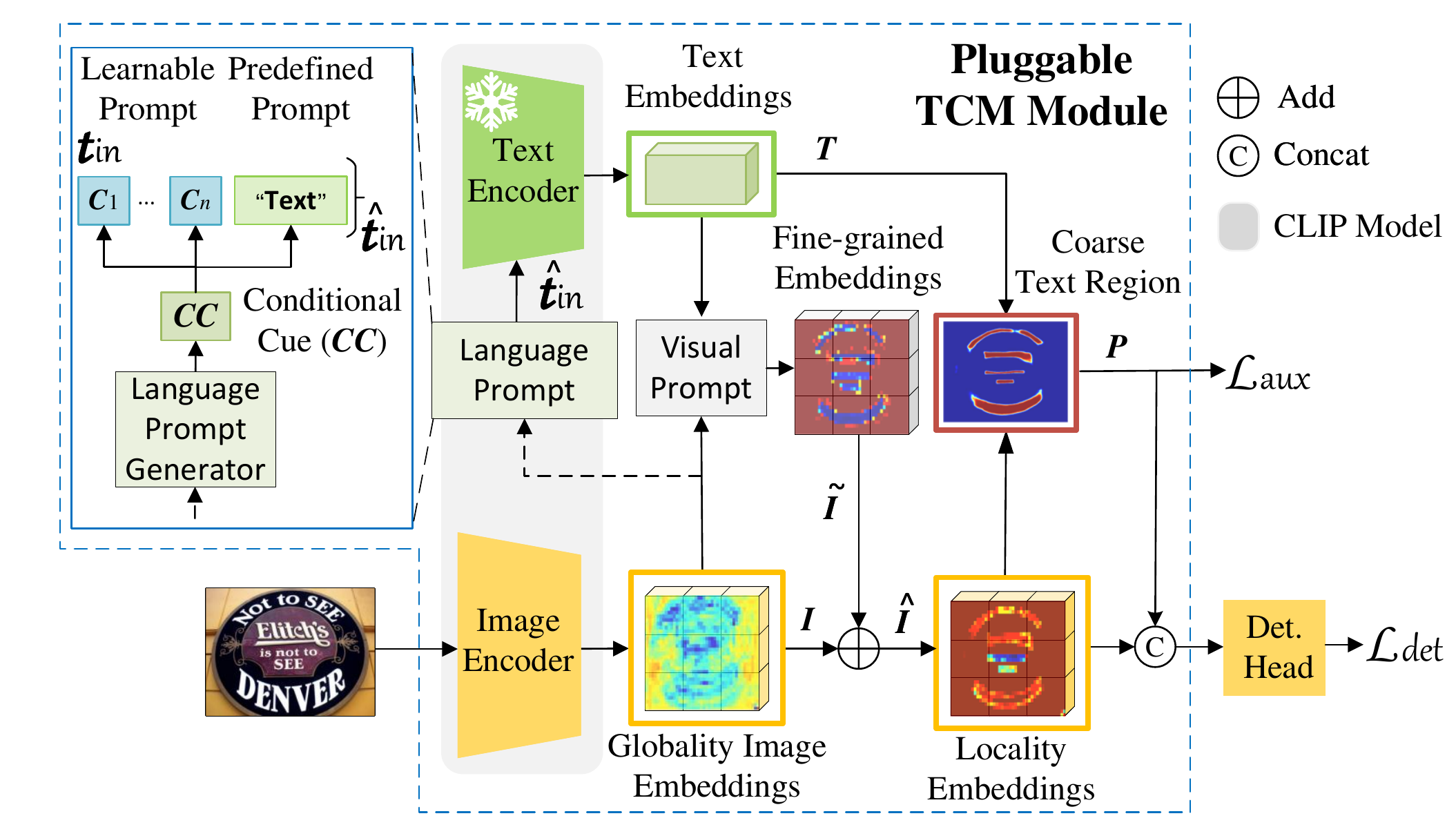}
\vspace{-1em}
\caption{The details of the TCM. The image encoder and text encoder are directly from the CLIP model. Det. Head short for detection head.}
\vspace{-1em}
\label{fig:method_plug_tcm}
\end{figure}

To turn the CLIP model into the scene text detector, we propose TCM, as shown in Fig.~\ref{fig:method_overall} and Fig.\ref{fig:method_plug_tcm}.
TCM is a pluggable module that can be directly applied to enhance the existing scene text detectors. It extracts the image and text embeddings from the image encoder and text encoder of CLIP model, respectively. 
We then design a cross-modal interaction mechanism through visual prompt learning 
to recover the locality feature from the image encoder of CLIP, which can capture fine-grained information to respond to the coarse text region for the subsequent matching between text instance and language.
For better steering the pretrained knowledge, we introduce a language prompt generator to generate conditional cue for each image and design a visual prompt generator that learns image prompts for adapting the frozen clip text encoder for the text detection task. The TCM can be directly applicable to broader text detection methods only with some minor modifications.
In addition, we design an instance-language matching method to align the image embedding and text embedding, which encourages the image encoder to explicitly refine text regions from cross-modal visual-language priors.

\paragraph{Image Encoder.} 
We use the pretrained ResNet50~\cite{He2016DeepRL} of CLIP as the image encoder, which produces an embedding vector for every input pixel.
Given the input image $ \bm{I}' \in \mathbb{R}^{ H \times W \times 3} $, image encoder outputs image embedding $\bm{I} \in \mathbb{R}^{\tilde{H} \times \tilde{W} \times C}$, where $\tilde{H} = \frac{H}{s}$, $\tilde{W} = \frac{W}{s}$, and $C$ is the image embedding  dimension ($C$ is set to 1024) and $s$ is the downsampling ratio (s is empirically set to 32), %
which can be expressed as:
\begin{equation} \label{eq:image_encoder}
\bm{I} =  \operatorname{ImageEncoder}(\bm{I}')\,.
\end{equation}

\paragraph{Text Encoder.} The text encoder takes input a number of of $K$ classes prompt and embeds it into a continuous vector space $\mathbb{R}^C$, producing text embeddings $ \bm{T} = \{\bm{t}_1,\ldots,\bm{t}_K\} \in \mathbb{R}^{K \times C}$ as outputs of the text encoder, where $ \bm{t}_i \in \mathbb{R}^C$. Specifically, we leverage the frozen pretrained text encoder of CLIP throughout as the text encoder can provide language knowledge prior for text detection. $K$ is set to 1 because there is only one text class in text detection task. Different from the original model that uses templates like ``a photo of a [CLS].'', we predefine discrete language prompt as ``\emph{Text}''. Then, a part of the text encoder input $\bm{t}_{in}'$ is defined as follows:
\begin{equation} \label{eq:image_text_encoder_input}
\bm{t}_{in}' =  \operatorname{WordEmbedding}( \rm Text) \in \mathbb{R}^{D}\,,
\end{equation}
where $\operatorname{WordEmbedding}(\cdot)$ denotes word embedding for predefined prompt ``Text'' class. $D$ is the word embedding dimension and set to 512.

Inspired by CoOp~\cite{Zhou2021LearningTP, Zhou2022ConditionalPL}, we also add learnable prompt $\{\bm{c}_1,\ldots,\bm{c}_n\}$ to learn robust transferability of text embedding for facilitating zero-shot transfer of CLIP model,  where $n$ is the number of learnable prompt, which is set to 4 by default, and $\bm{c}_i \in \mathcal{R}^D$. Thus, the input $\bm{t}_{in}$ of the text encoder is as follows: 
\begin{equation} \label{eq:t_in}
\bm{t}_{in} = [\bm{c}_1,\ldots,\bm{c}_n, \bm{t}_{in}'] \in \mathbb{R}^{(n+1) \times D}\,.
\end{equation}
 The text encoder takes $\bm{t}_{in}$ as input and generates text embedding $ \bm{T} = \{\bm{t}_1\} \in \mathbb{R}^{C}$, and $ \bm{T}$ is donated by $\bm{t}_{out} \in \mathcal{R}^C$ for simplification:
 \begin{equation} \label{eq:text_encoder}
 \bm{t}_{out} = \operatorname{TextEncoder}( \bm{t}_{in}) \in \mathbb{R}^{C}\,.
\end{equation}

\paragraph{Language Prompt Generator.}
Although the predefined prompt and learnable prompt are effective for steering the CLIP model, it may suffer from limited few-shot or generalization ability to open-ended scenarios where the testing text instance is out-of-distribution from the training images. 
To this end, we present a language prompt generator to generate a feature vector, termed as conditional cue ($\bm{cc}$). For each image,
the $\bm{cc}$ is then combined with the input of the text encoder $\bm{t}_{in}$, formulated as follows:
\begin{equation} \label{eq:text_gen}
\hat{\bm{t}}_{in} = \bm{cc} + \bm{t}_{in} \in \mathbb{R}^{(n+1) \times D}\,,
\end{equation}
where $\hat{\bm{t}}_{in}$ is the new prompt input of the text encoder conditioned on the input image, and we replace $\bm{t}_{in}$ with $\hat{\bm{t}}_{in}$ in Eq.~\ref{eq:text_encoder}.

In practice, the language prompt generator is built with a two-layer feed-forward network, which is applied to generate conditional cue ($\bm{cc}$) from the globality image embedding $\bm{I}$. This consists of two layer normalization followed by linear transformations, with a ReLU activation in between, which is formulated as follows:
\begin{equation} \label{eq:language_prompt_generator}
\bm{cc} = \operatorname{LN}(\sigma(\operatorname{LN}(\bm{\bar{I}})\bm{W}_1+\bm{b}_1)) \bm{W}_2+\bm{b}_2 \in \mathbb{R}^{D}\,,
\end{equation}
where $\bm{\bar{I}} \in \mathbb{R^{C}}$ is the global image-level feature generated from image embedding $\bm{I}$ by the same global attention pooling layer as in CLIP. $\bm{W}_1 \in \mathbb{R}^{ C \times C}$,  $\bm{W}_2 \in \mathbb{R}^{ C \times D}$, $\bm{b}_1 \in \mathbb{R}^{ C }$, $\bm{b}_2 \in \mathbb{R}^{ D }$, and we broadcast $\bm{cc}$ with $\bm{t}_{in}$ to get $\hat{\bm{t}}_{in}$ in Eq.~\ref{eq:text_gen}.

\paragraph{Visual Prompt Generator.}
We design a visual prompt generator to adaptively propagate fine-grained semantic information from textual features to visual features. 
Formally, we use the cross-attention mechanism in Transformer~\cite{Vaswani2017AttentionIA} to model the interactions between image embedding ($\bm{Q}$) and text embedding ($\bm{K}$, $\bm{V}$). The visual prompt $\tilde{\bm{I}}$ is then learned for transferring the information prior from image-level to text instance-level, which is defined as:
 \begin{equation} \label{eq:vis_prompt_gen}
 \tilde{\bm{I}} = \operatorname{TDec}( Q= \bm{I}, K= \bm{t}_{out}, V= \bm{t}_{out} ) \in \mathbb{R}^{ \tilde{H} \times \tilde{W} \times C}\,,
\end{equation}
where TDec denotes the Transformer Decoder.

Based on the conditional visual prompt, the original image embedding $\bm{I}$ is equipped with  
$\tilde{\bm{I}}$ to produce the prompted text-aware locality embeddings $\hat{\bm{I}}$ used for instance-language matching~(Eq.~\ref{eq:updated_pixel_text_matching}) and downstream detection head:
\begin{equation} \label{eq:vis_gen}
\hat{\bm{I}} = \bm{I} + \tilde{\bm{I}}.
\end{equation}

\paragraph{Instance-language Matching.}
Given the output of the text encoder and image encoder, we perform text instance-language matching alignment on text-aware locality image embedding $\hat{\bm{I}}$ and text embedding $\bm{t}_{out}$ by the dot product followed by sigmoid activation to get binary score map. The mixture of the generated conditional fine-grained embedding $\tilde{\bm{I}}$ and visual embedding $\bm{I}$ can allow text instance existing in visual features to be better matched with pretrained language knowledge in collaboration. The matching mechanism is formulated as follows:
\begin{equation} \label{eq:updated_pixel_text_matching}
\bm{P} = \operatorname{sigmoid}(  \hat{\bm{I}}\bm{t}_{out}^T / \tau ) \in \mathbb{R}^{\tilde{H} \times \tilde{W} \times 1},
\end{equation}
where $\bm{t}_{out}$ is text embedding because of only one text class in text detection scenarios, and $\bm{P}$ is the binary text segmentation map. The segmentation maps are supervised using the ground-truths as an auxiliary loss and concatenated by the prompted embedding $\hat{\bm{I}}$ for downstream text detection head to explicitly incorporate language priors for detection. During training, we minimize a binary cross-entropy loss between the segmentation map $\bm{P}$ and ground-truth, which is defined as follows:
\begin{equation} \label{eq:l_aux}
\mathcal{L}_{aux} = {\sum_i^{\tilde{H}}}\sum_j^{\tilde{W}} y_{ij}\log(P_{ij}) + (1-y_{ij})\log(1-P_{ij})\,,
\end{equation}
where $y_{ij}$ and $P_{ij}$ are the label and predicted probability of pixel $(i,j)$ belonging to the text instances, respectively. 

\paragraph{Optimization.}
The loss function $\mathcal{L}_{total}$ is the sum of detection loss $\mathcal{L}_{det}$ and auxiliary loss $\mathcal{L}_{aux}$, formulated as follows:
\begin{equation} \label{eq:total_loss}
\mathcal{L}_{total} = \mathcal{L}_{det} + \lambda \mathcal{L}_{aux} \,,
\end{equation}
where $\lambda$ is a trade-off hyper-parameters and set to 1 in this paper. $\mathcal{L}_{det}$ depends on downstream text detection method including segmentation and regression categories. In the inference period, we use the output of the detection head as the final result.

\section{Experiments}
\label{sec:experiments}
We conduct four sets of experiments to validate TCM. 
Our first set of experiment examines how TCM can be incorporated into existing text detectors to achieve consistent performance improvements.
Next, we demonstrate the few-shot training capability and generalization ability by incorporating the TCM method.
In the third set of experiments, we compare our method with previous pretraining methods. Finally, we provide thorough experiments to evaluate the sensitivity \wrt the proposed designs.

\paragraph{Datasets.} Our experiments are conducted on a number of commonly known scene text detection benchmarks including ICDAR2013 (IC13)~\cite{Karatzas2013ICDAR2R}, ICDAR2015 (IC15)~\cite{karatzas2015icdar}, MSRA-TD500 (TD)~\cite{yao2012detecting}, CTW1500 (CTW)~\cite{liu2019curved}, Total-Text (TT)~\cite{ch2019total}, ArT~\cite{chng2019icdar2019}, MLT17~\cite{Nayef2017ICDAR2017RR}, and MLT19~\cite{nayef2019icdar2019}.
More details of the datasets refer to appendix.

\paragraph{Evaluation Metric.} We use intersection over union (IoU) to determine whether the model correctly detects the region of text, and we calculate precision (P), recall (R), and F-measure (F) for comparison following common practice~\cite{Karatzas2013ICDAR2R}. For fair comparisons, text regions labeled with either ``do not care'' or ``\#\#\#'' will be ignored in all datasets during training and testing.

\paragraph{Implementation Details.}
For text detection tasks, we experiment with the popular text detection methods including DBNet (DB)~\cite{Liao2020RealtimeST}\footnote{\url{https://github.com/MhLiao/DB}}, PAN~\cite{Wang2019EfficientAA}\footnote{\url{https://github.com/whai362/pan_pp.pytorch}}, and FCENet (FCE)~\cite{zhu2021fourier}\footnote{\url{https://github.com/open-mmlab/mmocr/tree/main/configs/textdet/fcenet}} to evaluate TCM. For consistent settings with these methods, we train the detector using both SynthText and the real datasets. Specifically, the backbone is instantiated with the pretrained image encoder ResNet50~\cite{He2016DeepRL} of the CLIP unless specified. 
The visual prompt generator has 3 transformer decoder layers with 4 heads; transformer width is 256; and the feed-forward hidden dimension is set to 1024. We use the corresponding detection head of the DBNet, PAN, and FCENet to predict the final results. 
For testing few-shot learning of model, we directly train on the benchmark with different proportions of training data without pretraining and test it on the corresponding test data. 
For testing the generalization ability, we use the model trained on the corresponding source datasets and evaluating it on the target dataset that has dissimilar distribution. We consider two kinds of adaptation including synthtext-to-real and real-to-real, to validate the domain adaptation of the TCM. The ablation studies are conducted \wrt the predefined prompt, the learnable prompt, the language prompt generator, the visual prompt generator, and the different settings. The DBNet is used as baseline for TCM. 

\subsection{Cooperation with Existing Methods}
We report the text detection results of our TCM combined with three text detection methods on IC15, TD, and CTW in Table~\ref{tab:text_det_ic15}. Our method is +0.9\%, +1.7\%, and +1.9\% higher than the original FCENet, PAN, and DBNet, respectively, in terms of F-measure on IC15. TD and CTW also have similar consistent improvement. 
Note that the inference speed of our method is 18, 8.4, and 10 FPS evaluated on IC15, TD, and CTW datasets, respectively, with PAN, FCENet, and DBNet, remaining the high efficiency of the detector. %

  \begin{table}[htbp]
    \centering
    \normalsize
    \setlength\tabcolsep{2.4pt}

\newcommand{\tabincell}[2]{\begin{tabular}{@{}#1@{}}#2\end{tabular}}
{
\begin{tabularx}{1.0\linewidth}{c|cccccccccccccccccc}
    \toprule
     \multicolumn{1}{c}{\multirow{1}{*}{}} &
	\multirow{2}[2]{*}{Method} & 
	\multicolumn{2}{c}{IC15} &
	\multicolumn{2}{c}{TD} &
	\multicolumn{2}{c}{CTW} &
	\multirow{2}[1]{*}{FPS} 
	 \\
    \cmidrule(r){3-8}
	 \multicolumn{1}{c}{\multirow{1}{*}{}} & & F & $\Delta$ &  F & $\Delta$ & F & $\Delta$\\
    \midrule

\multirow{2}*{\rotatebox{90}{Reg.}} &	FCENet~\cite{zhu2021fourier} & 86.2 & - & 85.4$^\dagger$ & - & 85.5 & - & 11.5 \\
 &	TCM-FCENet  & \textbf{87.1} & \textbf{\greenp{+0.9}}   & \textbf{86.9} & \textbf{\greenp{+1.5}} & \textbf{85.9} & \textbf{\greenp{+0.4}} & 8.4 \\	
    \midrule
    
 \multirow{4}*{\rotatebox{90}{Seg.}} &	PAN~\cite{Wang2019EfficientAA}  & 82.9 & - & 84.1 & - & 83.7 & - & 36 \\
 &	TCM-PAN   & \textbf{84.6} & \textbf{\greenp{+1.7}} & \textbf{85.3} & \textbf{\greenp{+1.2}} & \textbf{84.3} & \textbf{\greenp{+0.6}} & 18 \\
    \cline{2-9}
    
 &	DBNet~\cite{Liao2020RealtimeST}  & 87.3 & - & 84.9 & - & 83.4 & - & 14.5 \\
&	TCM-DBNet & \textbf{89.2} & \textbf{\greenp{+1.9}} & \textbf{88.8} & \textbf{\greenp{+3.9}} & \textbf{84.9} & \textbf{\greenp{+1.5}} & 10  \\
    \bottomrule
\end{tabularx}
}

		\caption{
		Text detection results of cooperating with existing methods on IC15, TD, and CTW. $^\dagger$ indicates the results from~\cite{Zhan2019GADANGD}. Reg. and Seg. short for regression and segmentation methods, respectively. FPS are reported with ResNet50 backbone on a single V100.
		}
		\label{tab:text_det_ic15}
    \vspace{-1em}
	\end{table}

\begin{figure}[ht]
    \centering
    \subcaptionbox{CTW1500}{\includegraphics[width=3.5cm,height=2.3cm]{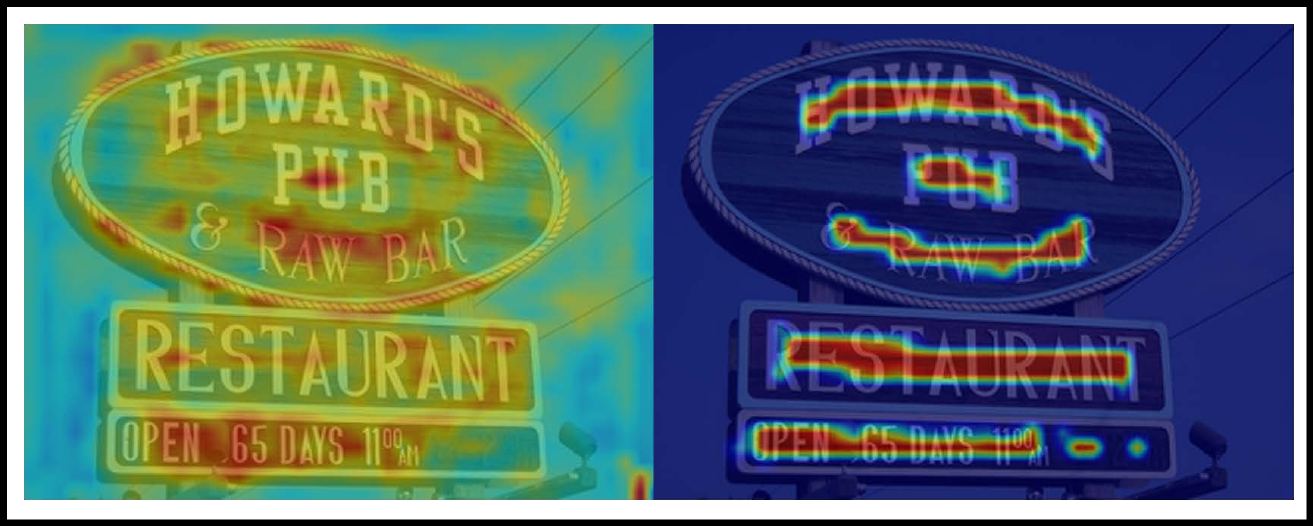}\label{fig:vis_ctw1500}}
    \subcaptionbox{MSRA-TD500}{\includegraphics[width=3.5cm,height=2.3cm]{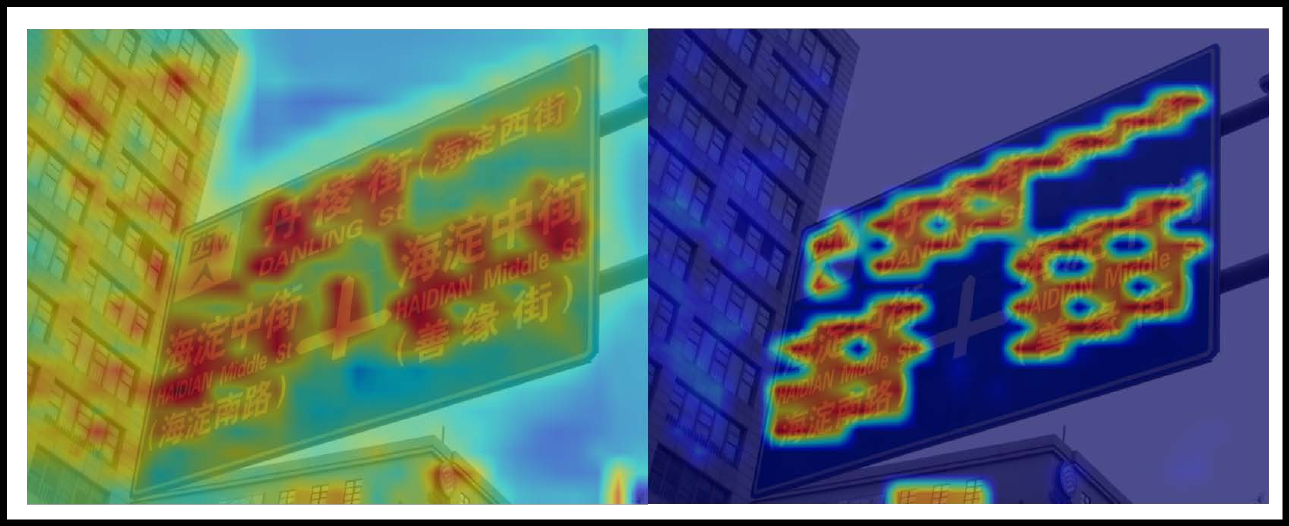}\label{fig:vis_roic13}}
    \vspace{-0.5em}
    \caption{Visualization results of our method. For each pair, the left is the image embedding $\bm{I}$, and the right is the generated visual prompt $\tilde{\bm{I}}$.  Best view in screen. More results can be found in appendix.}
    \label{fig:vp_results}
    \vspace{-1em}
\end{figure}

We visualize our method in Fig.~\ref{fig:vp_results}. It shows that the fine-grained features $\tilde{\bm{I}}$ containing text information is recovered from the global image embedding $\bm{I}$, demonstrating that TCM can identify text regions and provide this prior cues for downstream text detection.

\subsection{Few-shot Training Ability}
\begin{figure}[htbp]
    \centering
    \subcaptionbox{TD500}{\includegraphics[width=3.2cm,height=3.2cm]{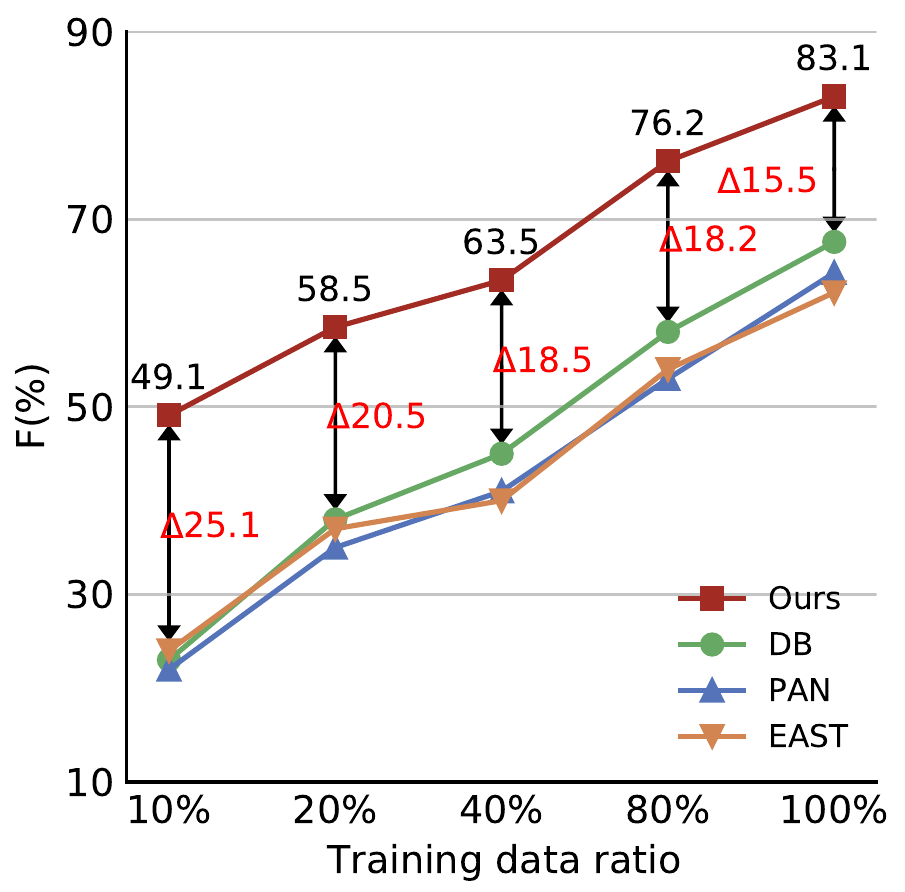}\label{fig:conditional_cue}}
    \subcaptionbox{ICDAR15}{\includegraphics[width=3.2cm,height=3.2cm]{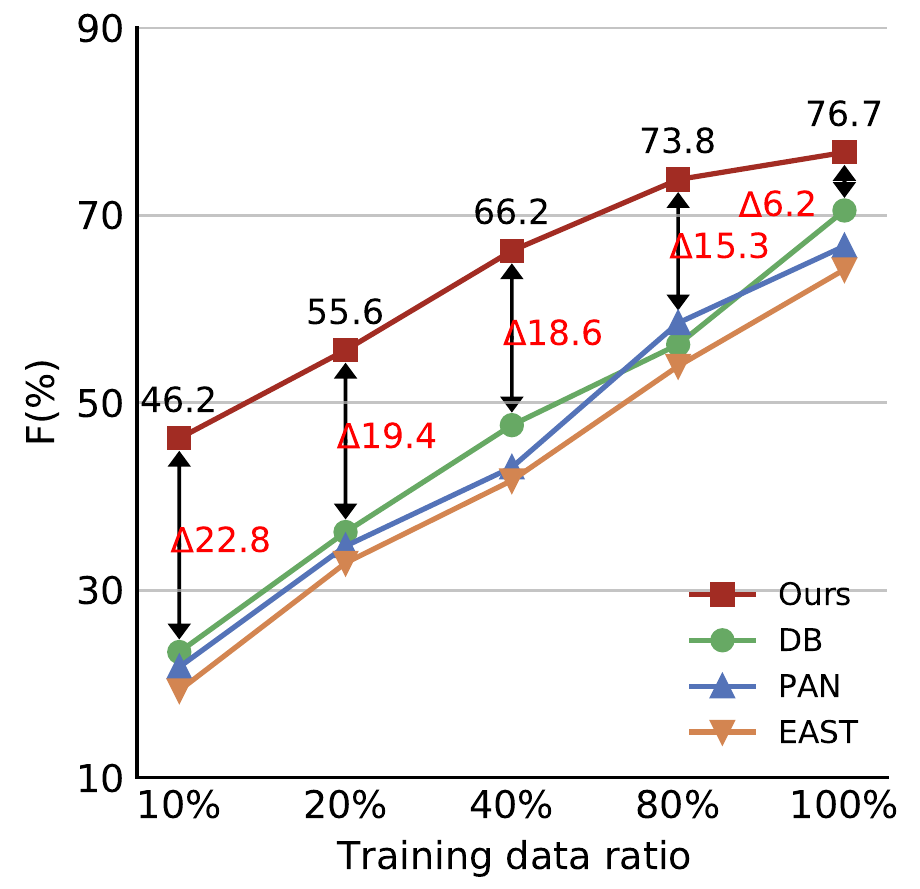}\label{fig:training_ratio_b}}
    \\
    \subcaptionbox{TotalText}{\includegraphics[width=3.2cm,height=3.2cm]{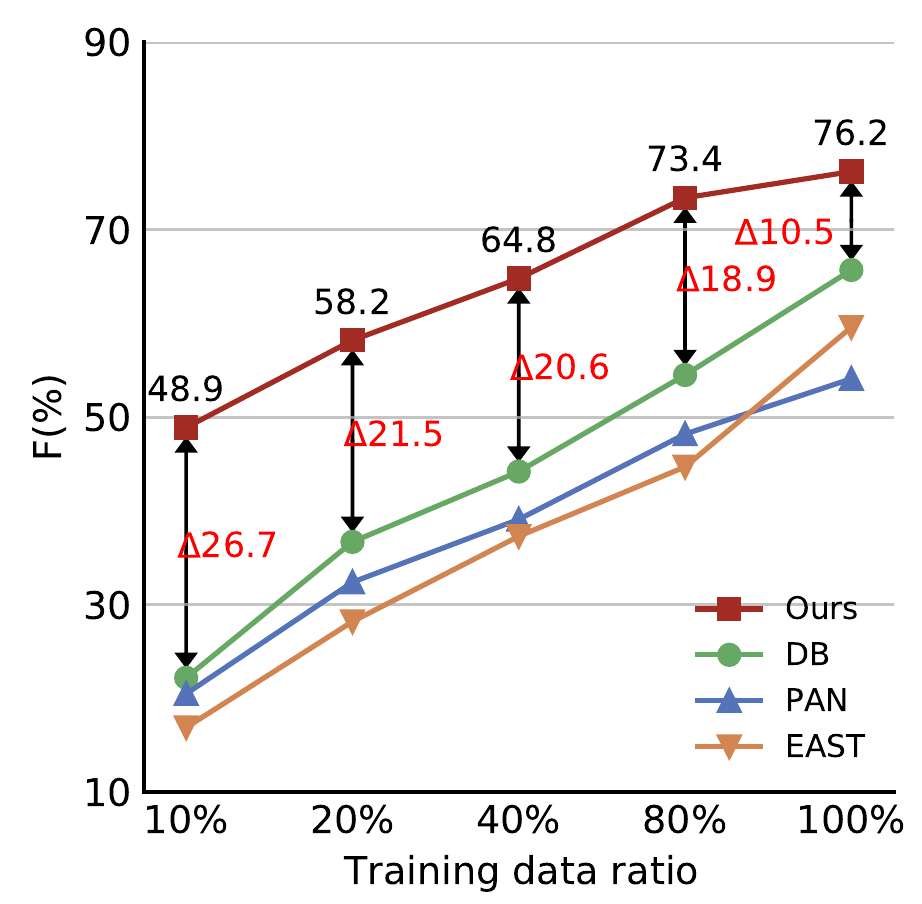}\label{fig:training_ratio_c}}
    \subcaptionbox{CTW1500}{\includegraphics[width=3.2cm,height=3.2cm]{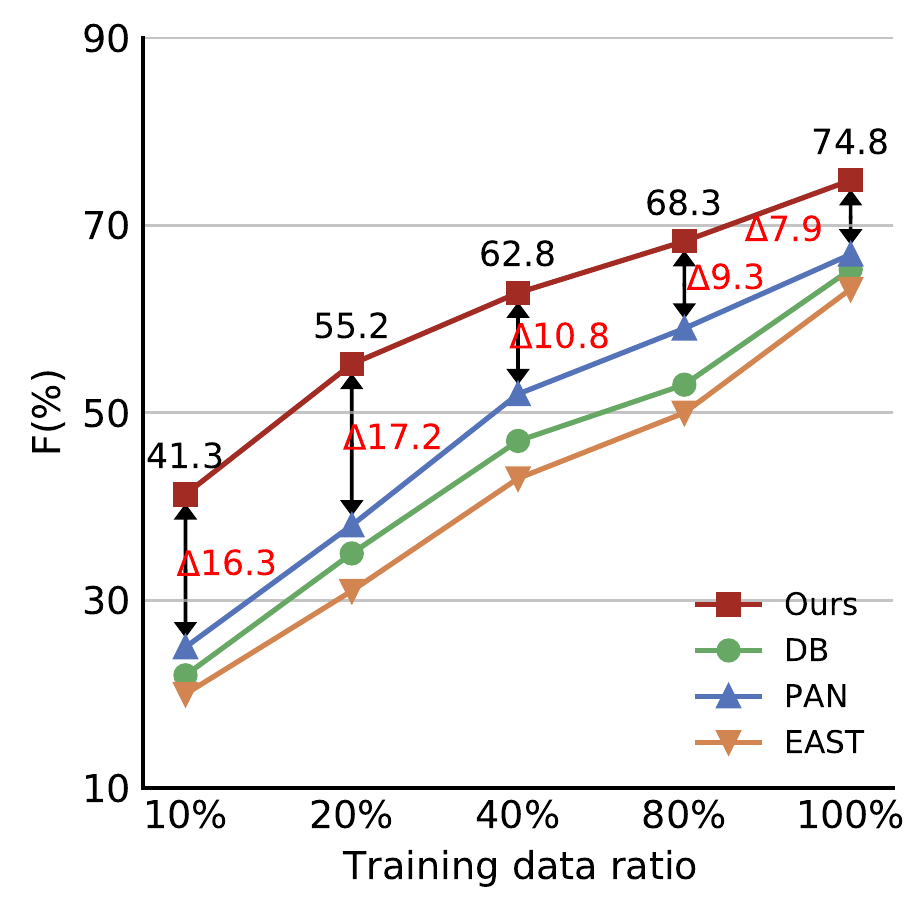}\label{fig:training_ratio_d}}
    \caption{Few-shot training ability with varying training data ratio. ``F'' represents F-measure. 
    }
    \label{fig:few_show_ability}
\end{figure}

To further verify the few-show training ability of our
method, we directly train our model on real datasets using various training data ratio without pretraining, and evaluate it on the corresponding 4 benchmarks.
As shown in Fig.~\ref{fig:few_show_ability}, our method shows robust on limited data and outperforms the three baseline methods including DB, PAN and EAST~\cite{Zhou2017EASTAE}. 
The results show that the TCM can capture the inherent characteristic of text via leveraging the pretrained vision and language knowledge of the zero-shot trained CLIP model.

\subsection{Generalization Ability}
\label{sec:exp:gn}
We conduct two types of experiments including synthtext-to-real adaptation and real-to-real adaptation, as shown in Table~\ref{tab:synth_to_real} and Table~\ref{tab:real_to_real}, respectively. 
From the tables, we can see that by plugging the TCM to DBNet, we significantly improve the performance by an average of 8.2\% in terms of F-measure for four different settings including synthtext-to-real and real-to-real, which further demonstrates the effectiveness of our method for domain adaptation.

	\begin{table}[t]
		\centering
		\renewcommand\arraystretch{1}
		\setlength{\tabcolsep}{2.0mm}
		\newcommand{\tabincell}[2]{\begin{tabular}{@{}#1@{}}#2\end{tabular}}
\begin{tabular}{lcc}
    \toprule
	\multirow{1}[0]{*}{Method} &
	\multicolumn{1}{c}{ST $\rightarrow$ IC13} &
	\multicolumn{1}{c}{ST $\rightarrow$ IC15}
	 \\
    \midrule
	EAST$^\dagger$~\cite{Zhou2017EASTAE} & 67.1 & 60.5   \\
	PAN~\cite{Wang2019EfficientAA} & - & 54.8   \\
	CCN~\cite{Xing2019ConvolutionalCN} & - & 65.1  \\
	ST3D~\cite{Liao2020SynthText3DSS} & 73.8 & 67.6  \\
	DBNet~\cite{Liao2020RealtimeST} & 71.7 & 64.0  \\
    \midrule
	TCM-DBNet & \textbf{79.6} & \textbf{76.7}  \\
    \bottomrule
\end{tabular}
		\caption{Synthtext-to-real adaptation. $^\dagger$ indicates the results from~\cite{Wu2020SynthetictoRealUD}. ST indicates SynthText. F-measure (\%) is reported.}
		\label{tab:synth_to_real}
    \vspace{-1.3em}
	\end{table}

	\begin{table}[ht]
		\centering
		\renewcommand\arraystretch{1}
		\setlength{\tabcolsep}{2.0mm}
	    \resizebox{\linewidth}{!}{
		\newcommand{\tabincell}[2]{\begin{tabular}{@{}#1@{}}#2\end{tabular}}
\begin{tabular}{lcc}
    \toprule
	\multirow{1}[0]{*}{Method} &
	\multicolumn{1}{c}{IC13 $\rightarrow$ IC15} &
	\multicolumn{1}{c}{IC13 $\rightarrow$ TD}
	 \\
    \midrule
	EAST$^\dagger$~\cite{Zhou2017EASTAE}  & 53.3 & 46.8  \\
	GD(AD)~\cite{Zhan2019GADANGD}  & 64.4  & 58.5  \\
	GD(10-AD)\cite{Zhan2019GADANGD}  & 69.4  & 62.1   \\
	CycleGAN~\cite{Zhu2017UnpairedIT}  & 57.2  & -   \\
	ST-GAN~\cite{Lin2018STGANST}  & 57.6  & -   \\
	CycleGAN+ST-GAN  & 60.8 & -   \\
	TST~\cite{Wu2020SynthetictoRealUD}  & 52.4  & -   \\
	DBNet~\cite{Liao2020RealtimeST}  & 63.9 & 53.8  \\
    \midrule
	TCM-DBNet & \textbf{71.9}  & \textbf{65.1}  \\
    \bottomrule
\end{tabular}

		}
		\caption{Real-to-real adaptation. $^\dagger$ indicates that the results are from~\cite{Zhan2019GADANGD}. %
		Note that the proposed method outperforms other methods. F-measure (\%) is reported.
		}
		\label{tab:real_to_real}
	    \vspace{-1.5em}
	\end{table}

\subsection{Comparison with Pretraining Methods}
\label{subset:pretrain}
The pretraining methods based on specifically designed pretext tasks has made effective progress in the field of text detection. In contrast to these efforts, TCM can turn the CLIP model directly into a scene text detector without pretraining process. The comparison results are shown in Table~\ref{tab:comap_pretraining}, from which we can see that without pretext tasks for pretraining, DB+TCM consistently outperforms previous methods including DB+STKM~\cite{Wan2021SelfattentionBT}, DB+VLPT~\cite{Song2022VisionLanguagePF}, and  DB+oCLIP~\cite{Xue2022LanguageMA}. Especially on IC15, our method outperforms previous state-of-the-art pretraining method by a large margin, with 89.4\% versus 86.5\% in terms of the F-measure.

\begin{table}[tb]
\centering
\normalsize
\setlength\tabcolsep{2.0pt}
{
\begin{tabularx}{1.0\linewidth}{l|lccccc}

\hline
                                                      & Methods   & Pretext task & IC15 & TT    & TD    & CTW   \\ \hline
\multirow{5}{*}{\rotatebox{90}{Convention}}                  & SegLink~\cite{Shi2017DetectingOT}   &     $\times$              & -    & -     & 77.0    & -     \\
                                                      & PSENet-1s~\cite{Li2019ShapeRT}    &          $\times$         & 85.7 & 80.9  & -     & 82.2  \\
                                                      & LOMO~\cite{Zhang2019LookMT}      &          $\times$         & 87.2 & 81.6  & -     & 78.4  \\
                                                      & MOST~\cite{He2021MOSTAM}      &          $\times$         & 88.2 & -     & 86.4  & -     \\
                                                      & Tang \etal\cite{Tang2022FewCB}      &            $\times$       & 89.1 & -     & 88.1  & -     \\ \hline
\multirow{4}{*}{\rotatebox{90}{VLP}} & DB+ST$^\dagger$     &      $\times$             & 85.4 & 84.7  & 84.9  & -     \\
                                                      & DB+STKM$^\dagger$~\cite{Wan2021SelfattentionBT}   &   \checkmark               & 86.1 & 85.5  & 85.9  & -     \\
                                                      & DB+VLPT$^\dagger$~\cite{Song2022VisionLanguagePF}   &       \checkmark           & 86.5 & 86.3  & 88.5  & -     \\
                                                      & DB+oCLIP*~\cite{Xue2022LanguageMA}  &       \checkmark           & -    & -     & -     & 84.4  \\\hline
                                                      & DB+TCM(Ours) &         $\times$          & \textbf{89.4} & 85.9 & \textbf{88.8} & \textbf{85.1} \\ \hline

\end{tabularx}}
\caption{Comparison with existing scene text pretraining techniques on DBNet (DB). $^\dagger$ indicates the results from~\cite{Song2022VisionLanguagePF}. ST and VLP denote SynthText pretraining and visual-language pretraining methods, respectively.  * stand for our reimplementation results. F-measure (\%) is reported. 
}
\label{tab:comap_pretraining}
\vspace{-0.6em}
\end{table}

\subsection{Ablation Studies}
\noindent\textbf{Pretrained CLIP Backbone.} First, we conduct experiments that we only replace the original backbone of the DBNet with the pretrained image encoder ResNet50 of the CLIP to quantify the performance variance of the backbones. As shown in Table~\ref{tab:abla_clip_on_db}, 
the original pretrained model of CLIP is insufficient for leveraging the visual-language knowledge of the CLIP. Therefore, it is necessary to use a proper method to excavate the knowledge of the CLIP model. 
    
    \begin{table}[htbp]
	
        \setlength\tabcolsep{1pt}
		\centering
		\renewcommand\arraystretch{1}

		\setlength{\tabcolsep}{2.0mm}
		\resizebox{\linewidth}{!}{
		\newcommand{\tabincell}[2]{\begin{tabular}{@{}#1@{}}#2\end{tabular}}
\begin{tabular}{rccccc}
    \toprule
	\multirow{1}[0]{*}{Method} & 
    \multirow{1}[0]{*}{\tabincell{c}{BB}} & 
	\multicolumn{1}{c}{IC15} &
	\multicolumn{1}{c}{TD} &
	\multicolumn{1}{c}{TT} &
	\multicolumn{1}{c}{CTW} 
	 \\
    \midrule
	DBNet & R50  & 87.3  & 84.9 & 84.7 & 83.4 \\
	DBNet & CR50 & 87.7 (\greenp{+0.4}) & 86.8 (\greenp{+1.9}) & 84.7 & 83.4 \\
    \bottomrule
\end{tabular}

		}
		\caption{Ablation study of the ResNet50 backbone on IC15, TD, TT, and CTW. BB indicates Backbone. R50 and CR50 represent the ResNet50 backbones of the DBNet and the CLIP, respectively. 
		F-measure (\%) is reported. %
		}
		\label{tab:abla_clip_on_db}
    \vspace{-0.2cm}
	\end{table}

    \begin{table}[htbp]
        \centering
        \normalsize
        \setlength\tabcolsep{3.8pt}

		\newcommand{\tabincell}[2]{\begin{tabular}{@{}#1@{}}#2\end{tabular}}
{
\begin{tabularx}{1.0\linewidth}{rcccccccc}
\hline
    \toprule
	\multirow{2}[2]{*}{Method} & \multirow{2}[2]{*}{PP} &
	\multirow{2}[2]{*}{LP} &
	\multirow{2}[2]{*}{\tabincell{c}{LG}} & 
    \multirow{2}[2]{*}{\tabincell{c}{VG}} & 
	\multicolumn{1}{c}{IC15} &
	\multicolumn{1}{c}{TD} &
	\multicolumn{1}{c}{TT} &
	\multicolumn{1}{c}{CTW} 
	 \\
    \cmidrule(r){6-9}
	& & & & & F & F & F & F \\
    \midrule

	BSL & $\times$  & $\times$  & $\times$  & $\times$ & 87.7 & 86.8 & 84.7 & 83.4  \\
    \midrule

	BSL+ & \checkmark  & $\times$  & $\times$  & $\times$ & 87.75  & 87.0 & 84.74 &  83.5  \\
	BSL+ & \checkmark  & 4  & $\times$  & $\times$ & 88.0  & 87.1 & 84.8 &  83.6   \\
    \midrule
	BSL+ & $\times$   & 4  & $\times$  & $\times$ & 87.8  & 87.7 &  85.1 &  83.9   \\
	BSL+ & $\times$   & 18  & $\times$  & $\times$ & 88.1  & 87.8 & 85.3 & 83.9   \\
	BSL+ & $\times$   & 32  & $\times$  & $\times$ & 88.4  &  88.2 & 85.4 &  84.5  \\
    \midrule
	BSL+ & \checkmark  & 4  & \checkmark  & $\times$ &  88.6  & 88.4  & 85.5 &  84.6  \\
	TCM & \checkmark  & 4  & \checkmark  & \checkmark  & 89.2  & 88.8  &  85.6 & 84.9  \\
	TCM & \checkmark  & 32  & \checkmark  & \checkmark  & \textbf{89.4}   & \textbf{88.8}   & \textbf{85.9} & \textbf{85.1}   \\
	$\Delta$ &  &   &  &    & \greenp{+1.7} &  \greenp{+2.0}  & \greenp{+1.2} & \greenp{+1.7}  \\
    \bottomrule
\end{tabularx}
}
		\caption{Ablation study of our proposed components on IC15, TD, TT and CTW. 
		``BSL'', ``PP'', ``LP'', ``LG'', and ``VG'' represent the baseline method DBNet, the predefined prompt, the learnable prompt, the language prompt generator, and the visual prompt generator, respectively. F (\%) represents F-measure. $\Delta $ represents the variance.
		}
		\label{tab:abla_ic15_td}
    \vspace{-1.5em}
	\end{table}
	
  	\begin{table*}[htbp]
		\centering
		\renewcommand\arraystretch{1}
		\setlength{\tabcolsep}{2.0mm}
		\newcommand{\tabincell}[2]{\begin{tabular}{@{}#1@{}}#2\end{tabular}}
\begin{tabular}{lcccccc}
    \toprule
	\multirow{1}[0]{*}{Method} & 
	\multicolumn{1}{c}{IC13 $\rightarrow$ IC15} &
	\multicolumn{1}{c}{IC13 $\rightarrow$ TD} &
	\multicolumn{1}{c}{IC15 $\rightarrow$ MLT17(en)} &
	\multicolumn{1}{c}{TT $\rightarrow$ ArT(-)} &
	\multicolumn{1}{c}{ST $\rightarrow$ IC13} &
	\multicolumn{1}{c}{ST $\rightarrow$ IC15} 
	 \\
    \midrule
	TCM  & 71.9 &  65.1  & 85.1 & 68.9 & 79.5 & 76.7 \\
    \midrule
	w/o VG   & 68.4 (\redp{-3.5}) & 59.4 (\redp{-5.7})  & 81.8 (\redp{-3.3})  & 59.1 (\redp{-9.8}) & 76.3 (\redp{-3.2}) &  72.6 (\redp{-4.1}) \\
	w/o LG   & 66.1 (\redp{-5.8}) & 56.8 (\redp{-8.3})  & 79.7 (\redp{-5.4})  & 57.8 (\redp{-11.1}) & 74.5 (\redp{-5.0}) & 68.2 (\redp{-8.5})  \\
w/o VG \& LG & 64.8 (\redp{-7.1})  & 55.7 (\redp{-9.4})  & 78.4 (\redp{-6.7})  & 54.2 (\redp{-14.7}) & 71.7 (\redp{-7.8}) &  63.9 (\redp{-12.8}) \\	
    \bottomrule
\end{tabular}

		\caption{
		Ablation study of the effect of LG and VG on generalization performance. F-measure (\%) is reported.
		}
		\label{tab:abla_da_tg_vg}
		\vspace{-1.5em}
	\end{table*}

\noindent\textbf{Ablation Study for the Predefined Prompt.} When using the predefined prompt, as illustrated in the second row of Table~\ref{tab:abla_ic15_td}, the performances are slightly improved on all four datasets (IC15, TD, TT, and CTW), with 0.05\%, 0.2\%, 0.04\%, and 0.1\% higher than the baseline method, respectively.

\noindent\textbf{Ablation Study for the Learnable Prompt.} Besides, results combing the learnable prompt with the predefined prompt on four datasets are provided in the third row of Table~\ref{tab:abla_ic15_td}. We notice that a consistent improvement can be achieved by adding the learnable prompt.
We also show the influence of using different numbers of the learnable prompt in row 4 to row 6 of Table~\ref{tab:abla_ic15_td}. We observe that as the value of the number of the learnable prompt increases, the performance increases gradually on all datasets. Compared to the value 4, the value 32 obtains obvious improvements on CTW, TD, and TT. We conjecture that this is because the larger number of the learnable prompt can better steer the pretrained text encoder knowledge which is useful for text detection. In the following experiments, the default number of the learnable prompt is set to 4 for simplicity.

\noindent\textbf{Ablation Study for the Language Prompt Generator.} Furthermore, we evaluate the performance of the proposed language prompt generator shown in 7$_{th}$ row  of Table~\ref{tab:abla_ic15_td}. 
With the help of the language prompt generator, we find that TCM achieves further improvements on all four datasets, especially on ICDAR2015, indicating that the conditional cue generated by the language prompt generator for each image can ensure better generalization over different types of datasets.

\noindent\textbf{Ablation Study for the Visual Prompt Generator.} Finally, combining the proposed visual prompt generator with the above other components, the improvement of F-measure is better than the baseline on all four datasets, with larger margins of 1.7\% and 2.0\% on IC15 and TD, respectively.
The reason for this obvious complementary phenomenon is that the visual prompt generator can propagate fine-grained visual semantic information from textual features to visual features. 
Besides, the prompted locality image embedding generated by the visual prompt generator can guide the model to obtain more accurate text instance-level visual representations, which boosts the ability of instance-language matching and generates a precise segmentation score map that is useful for downstream detection head.

\noindent\textbf{Ablation Study for the VG and LG on Generalization Performance.} As described in Table~\ref{tab:abla_da_tg_vg}, removing the VG and LG elements from TCM dramatically deteriorates the generalization performance, which further indicates the effectiveness of the VG and LG. 

\noindent\textbf{Ablation Study for Image Encoder and Text Encoder.} We have investigated how the quality of the frozen text encoder and image encoder affects the performance via adjusting the corresponding learning rate (LR) factor. The experimental results of TCM-DBNet on the TD500 dataset are shown in Table~\ref{tab:appendix_expl_ie_te}. The results show that using a lower learning rate for both encoders and fixing the text encoder is the optimal setting for training the whole model. Note that we observe performance degradation when directly using $1.0\times$ learning rate for both encoders, which suggests the frozen text encoder can stabilize the training process. The cores of the  architecture, including the language prompt generator and visual prompt generator, are designed to better steer knowledge of the pretrained CLIP. Appropriate design of the network architecture and the use of the pretrained CLIP are complementary.

	\begin{table}[ht]
		\centering
		\renewcommand\arraystretch{1}

		\setlength{\tabcolsep}{2.0mm}

		\begin{tabular}{llll}
    \toprule
                           & Image encoder & Text encoder & F (\%) \\ 
       \midrule
\multirow{4}{*}{LR Factor} & 0.1           & 0.0          & \textbf{88.7}  \\
                           & 0.1           & 0.1          & 87.8  \\
                           & 0.1           & 1.0          & 87.1  \\
                           & 1.0           & 1.0          & 86.3  \\ 
       \bottomrule
\end{tabular}

		\caption{Ablation study of exploration on image encoder and text encoder. 
		``LR'' represents the learning rate.
		}
		\label{tab:appendix_expl_ie_te}
	\end{table}

\begin{figure}[htbp]
\centering

\includegraphics[width=0.47\textwidth]{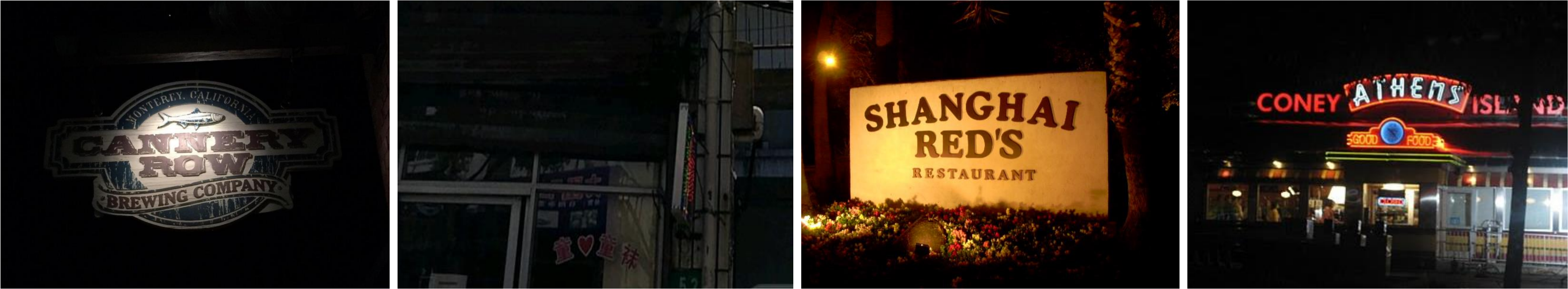}
\caption{The examples of our constructed NightTime-ArT.}
\vspace{-1em}
\label{fig:night_image}
\vspace{-0.1em}
\end{figure}

\noindent\textbf{Ablation Study for Different Amount of Data.}
To further explore whether the TCM can learn the additional knowledge which is hard to be obtained from increasing data, we have trained the model on a large-scale public joint data including IC13, IC15, TD, CTW, TT, and MLT17, with total 13,784 image, and testing it on a NightTime-ArT data (326 images) carefully collected from ArT. The nighttime examples of ArT are shown in Fig.~\ref{fig:night_image}.
Results are shown in Table~\ref{tab:appendix_more_data}. The results show that even with the addition of large amounts of training data, existing methods still show limitation to the nighttime data that is obviously out-of-distribution from the training set. However, TCM can still perform robust in such case, indicating its irreplaceable potential generalization ability.

\begin{table}[h]
\begin{tabular}{llll}
\toprule
Method         & Training Data       & Testing Data             & F (\%)      \\ 
\midrule
FCENet         & Joint data & NightTime-ArT  & 55.2        \\
DBNet          & Joint data & NightTime-ArT   & 52.8        \\ \midrule
TCM-DBNet & Joint data  & NightTime-ArT  & \textbf{70.2}  \\ \bottomrule
\end{tabular}
\caption{Ablation study of exploration on large amounts of training data.
}		
\label{tab:appendix_more_data}
		\vspace{-0.2cm}
\end{table}

\noindent\textbf{Ablation Study for the Parameters Comparison.}
For a fair comparison, we have increased the parameters of DBNet by replacing the backbone with a larger ResNet and then conduct experiments on TD500 dataset. Trainable parameters and FLOPs are calculated with an input size 1280$\times$800. Results are shown in Table~\ref{tab:appendix_fair_db}.
The results show that TCM-DBNet has better performance than DBNet with less model size and computation overhead, demonstrating its effectiveness for scene text detection.

	\begin{table}[ht]
		\centering
		\renewcommand\arraystretch{1}

		\setlength{\tabcolsep}{2.0mm}
		
\begin{tabular}{lllll}
\toprule
Method    & Backbone & Params & FLOPs & F (\%) \\ \midrule
DBNet     & R50      &    26 (M)  &   98 (G)  & 84.9   \\
DBNet     & R101     &    46 (M)  &   139 (G) & 85.9     \\
DBNet     & R152     &    62 (M)  &  180 (G)  & 87.3      \\ \midrule
TCM-DBNet & R50      &    50 (M)  &  156 (G)  & \textbf{88.7}  \\ \bottomrule
\end{tabular}
		\caption{Ablation study of the parameters comparison with DBNet.
		}
		\label{tab:appendix_fair_db}
		\vspace{-1.5em}
	\end{table}

\noindent\textbf{Ablation Study for the Auxiliary Loss.}
We further compare the results of with and without auxiliary loss on TD500 dataset, as shown in Table~\ref{tab:appendix_aux_loss}. We see that using auxiliary loss achieves higher performance. The results indicate auxiliary loss is beneficial to train the model via imposing constraints on instance-language matching score map. In addition, the improvement of the performance suggests that it might help the image encoder of pretrained CLIP to perceive locality text region effectively.

	\begin{table}[ht]
		\centering
		\renewcommand\arraystretch{1}

		\setlength{\tabcolsep}{2.0mm}
		
\begin{tabular}{ll}
\toprule
Model                   & F (\%) \\ \midrule
TCM-DBNet with auxiliary loss & \textbf{88.7}  \\
TCM-DBNet w/o auxiliary loss  & 85.1     \\ \bottomrule
\end{tabular}
		\caption{Ablation study of the auxiliary Loss.
		}
		\label{tab:appendix_aux_loss}
		\vspace{-0.5em}
	\end{table}

\begin{figure}[htbp]
\centering
\includegraphics[width=0.43\textwidth]{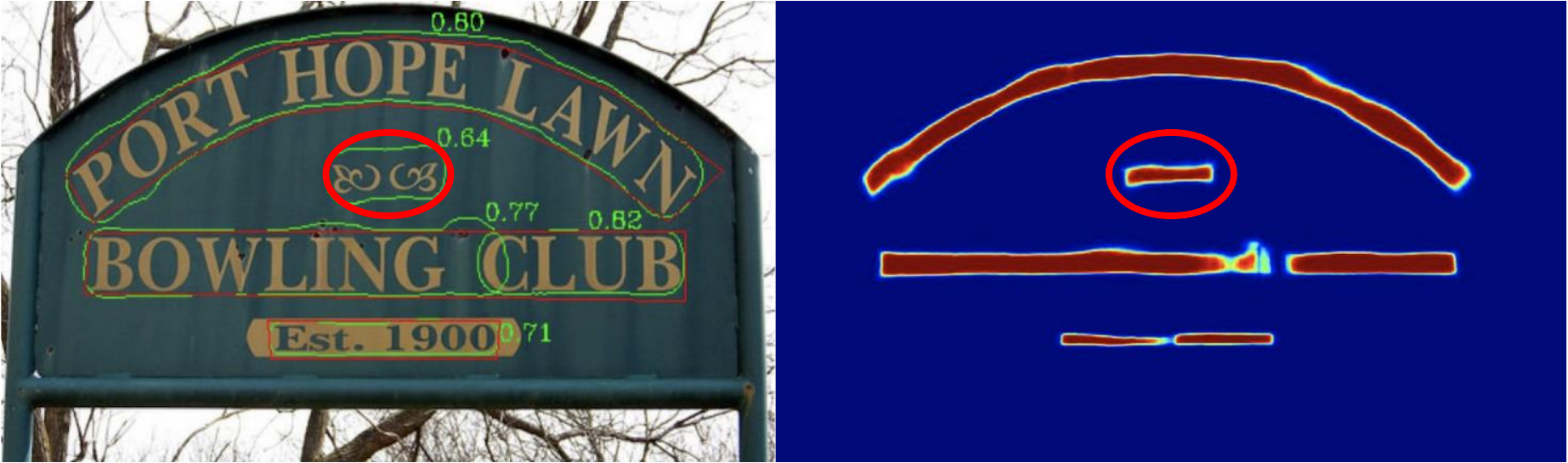}
\vspace{-0.5em}
\caption{Failure cases. Red circle means false positive region. }
\vspace{-1em}
\label{fig:failure}
\end{figure}

\section{Discussion of Failure Cases}
\label{sec:dis}
There are some insightful failed cases as shown in Figure~\ref{fig:failure}. The instance-language matching score map generates false positive region that is very similar to the characteristics of text, as shown in the region of the red circle in Fig.~\ref{fig:failure}, which will be considered as noise. Therefore, it is necessary that the downstream text detection head can further refine this initial score map instead of directly using the score map of instance-language matching as the final results. We leave this problem as future work to alleviate the false positive score map of instance-language matching.

\section{Conclusion}\label{sec:conclusion}
This paper proposes the TCM, which can directly excavate the prior knowledge from the CLIP model into a scene text detector without pretraining process. 
Such a new text detection paradigm reveals the importance of using visual-language prior for seeking information from the zero-shot off-the-rack model, and thus guiding the text detector adapting to small-scale data, divergent data distribution, and complicated scenes, without relying on carefully-designed pretraining tasks.
Experiments comprehensively demonstrate the effectiveness of our method. It is worth mentioning that we also construct a NightTime-ArT dataset to further demonstrate that the TCM can steer useful prior knowledge from the CLIP model.
As the CLIP model is an inborn-friendly framework for text, extension of TCM to scene text spotting is also a promising direction for future work.

\noindent\textbf{Acknowledgements} This work was supported by the National Natural Science Foundation of China (No.62225603, No.6220073278, No.62206103), and the National Key Research and Development Program (No.2022YFC3301703, No.2022YFC2305102).

{\small
\bibliographystyle{ieee_fullname}
\bibliography{references}

\begin{thebibliography}{10}\itemsep=-1pt

\bibitem{Baek2019CharacterRA}
Youngmin Baek, Bado Lee, Dongyoon Han, Sangdoo Yun, and Hwalsuk Lee.
\newblock Character region awareness for text detection.
\newblock In {\em 2019 IEEE/CVF Conference on Computer Vision and Pattern
  Recognition (CVPR)}, pages 9357--9366, 2019.

\bibitem{chng2019icdar2019}
Chee-Kheng Chng, Yuliang Liu, Yipeng Sun, Chun~Chet Ng, Canjie Luo, Zihan Ni,
  ChuanMing Fang, Shuaitao Zhang, Junyu Han, Errui Ding, et~al.
\newblock {{ICDAR2019} Robust Reading Challenge on Arbitrary-Shaped Text
  (RRC-ArT)}.
\newblock In {\em ICDAR}, pages 1571--1576, 2019.

\bibitem{ch2019total}
Chee-Kheng Ch’ng, Chee~Seng Chan, and Cheng-Lin Liu.
\newblock Total-text: toward orientation robustness in scene text detection.
\newblock {\em IJDAR}, pages 1--22, 2019.

\bibitem{goh2021multimodal}
Gabriel Goh, Nick Cammarata, Chelsea Voss, Shan Carter, Michael Petrov, Ludwig
  Schubert, Alec Radford, and Chris Olah.
\newblock Multimodal neurons in artificial neural networks.
\newblock {\em Distill}, 6(3):e30, 2021.

\bibitem{Gu2022OpenvocabularyOD}
Xiuye Gu, Tsung-Yi Lin, Weicheng Kuo, and Yin Cui.
\newblock Open-vocabulary object detection via vision and language knowledge
  distillation.
\newblock In {\em ICLR}, pages 1--20, 2022.

\bibitem{gupta2016synthetic}
Ankush Gupta, Andrea Vedaldi, and Andrew Zisserman.
\newblock Synthetic data for text localisation in natural images.
\newblock In {\em Proceedings of the IEEE conference on computer vision and
  pattern recognition}, pages 2315--2324, 2016.

\bibitem{He2016DeepRL}
Kaiming He, X. Zhang, Shaoqing Ren, and Jian Sun.
\newblock Deep residual learning for image recognition.
\newblock In {\em 2016 IEEE Conference on Computer Vision and Pattern
  Recognition (CVPR)}, pages 770--778, 2016.

\bibitem{He2021MOSTAM}
Minghang He, Minghui Liao, Zhibo Yang, Humen Zhong, Jun Tang, Wenqing Cheng,
  Cong Yao, Yongpan Wang, and Xiang Bai.
\newblock Most: A multi-oriented scene text detector with localization
  refinement.
\newblock In {\em 2021 IEEE/CVF Conference on Computer Vision and Pattern
  Recognition (CVPR)}, pages 8809--8818, 2021.

\bibitem{He2017SingleST}
Pan He, Weilin Huang, Tong He, Qile Zhu, Yu Qiao, and Xiaolin Li.
\newblock Single shot text detector with regional attention.
\newblock In {\em 2017 IEEE International Conference on Computer Vision
  (ICCV)}, pages 3066--3074, 2017.

\bibitem{He2017DeepDR}
Wenhao He, Xu-Yao Zhang, Fei Yin, and Cheng-Lin Liu.
\newblock Deep direct regression for multi-oriented scene text detection.
\newblock In {\em 2017 IEEE International Conference on Computer Vision
  (ICCV)}, pages 745--753, 2017.

\bibitem{karatzas2015icdar}
D. Karatzas, L. Gomez-Bigorda, et~al.
\newblock {ICDAR 2015 competition on robust reading}.
\newblock In {\em ICDAR}, pages 1156--1160, 2015.

\bibitem{Karatzas2013ICDAR2R}
Dimosthenis Karatzas, Faisal Shafait, Seiichi Uchida, M. Iwamura,
  Llu{\'i}s~G{\'o}mez i Bigorda, Sergi~Robles Mestre, Joan~Mas Romeu,
  David~Fern{\'a}ndez Mota, Jon Almaz{\'a}n, and Llu{\'i}s-Pere de~las Heras.
\newblock Icdar 2013 robust reading competition.
\newblock In {\em 2013 12th International Conference on Document Analysis and
  Recognition}, pages 1484--1493, 2013.

\bibitem{Li2022LanguagedrivenSS}
Boyi Li, Kilian~Q. Weinberger, Serge~J. Belongie, Vladlen Koltun, and Ren{\'e}
  Ranftl.
\newblock Language-driven semantic segmentation.
\newblock In {\em ICLR}, 2022.

\bibitem{Li2019ShapeRT}
Xiang Li, Wenhai Wang, Wenbo Hou, Ruo-Ze Liu, Tong Lu, and Jian Yang.
\newblock Shape robust text detection with progressive scale expansion network.
\newblock In {\em 2019 IEEE/CVF Conference on Computer Vision and Pattern
  Recognition (CVPR)}, pages 9328--9337, 2019.

\bibitem{Liao2017TextBoxesAF}
Minghui Liao, Baoguang Shi, Xiang Bai, Xinggang Wang, and Wenyu Liu.
\newblock Textboxes: A fast text detector with a single deep neural network.
\newblock In {\em AAAI}, pages 4161--4167, 2017.

\bibitem{Liao2020SynthText3DSS}
Minghui Liao, Boyu Song, Minghang He, Shangbang Long, Cong Yao, and Xiang Bai.
\newblock Synthtext3d: synthesizing scene text images from 3d virtual worlds.
\newblock {\em Science China Information Sciences}, 63:1--14, 2020.

\bibitem{Liao2019RealTimeST}
Minghui Liao, Zhaoyi Wan, Cong Yao, Kai Chen, and Xiang Bai.
\newblock Real-time scene text detection with differentiable binarization and
  adaptive scale fusion.
\newblock {\em IEEE Transactions on Pattern Analysis and Machine Intelligence},
  45:919--931, 2019.

\bibitem{Liao2020RealtimeST}
Minghui Liao, Zhaoyi Wan, Cong Yao, Kai Chen, and Xiang Bai.
\newblock Real-time scene text detection with differentiable binarization.
\newblock In {\em AAAI}, pages 11474--11481, 2020.

\bibitem{Lin2018STGANST}
Chen-Hsuan Lin, Ersin Yumer, Oliver Wang, Eli Shechtman, and Simon Lucey.
\newblock St-gan: Spatial transformer generative adversarial networks for image
  compositing.
\newblock In {\em 2018 IEEE/CVF Conference on Computer Vision and Pattern
  Recognition}, pages 9455--9464, 2018.

\bibitem{liu2019curved}
Yuliang Liu, Lianwen Jin, Shuaitao Zhang, Canjie Luo, and Sheng Zhang.
\newblock {Curved scene text detection via transverse and longitudinal sequence
  connection}.
\newblock {\em Pattern Recognition}, 90:337--345, 2019.

\bibitem{Long2020SceneTD}
Shangbang Long, Xin He, and Cong Yao.
\newblock Scene text detection and recognition: The deep learning era.
\newblock {\em International Journal of Computer Vision}, 129:161--184, 2020.

\bibitem{Long2022TowardsEU}
Shangbang Long, Siyang Qin, Dmitry Panteleev, A. Bissacco, Yasuhisa Fujii, and
  Michalis Raptis.
\newblock Towards end-to-end unified scene text detection and layout analysis.
\newblock {\em 2022 IEEE/CVF Conference on Computer Vision and Pattern
  Recognition (CVPR)}, pages 1039--1049, 2022.

\bibitem{Long2018TextSnakeAF}
Shangbang Long, Jiaqiang Ruan, W. Zhang, Xin He, Wenhao Wu, and Cong Yao.
\newblock Textsnake: A flexible representation for detecting text of arbitrary
  shapes.
\newblock In {\em ECCV}, pages 1--17, 2018.

\bibitem{nayef2019icdar2019}
Nibal Nayef, Yash Patel, Michal Busta, Pinaki~Nath Chowdhury, Dimosthenis
  Karatzas, Wafa Khlif, Jiri Matas, Umapada Pal, Jean-Christophe Burie,
  Cheng-lin Liu, et~al.
\newblock {{ICDAR2019} Robust Reading Challenge on Multi-lingual Scene Text
  Detection and Recognition--RRC-MLT-2019}.
\newblock In {\em ICDAR}, pages 1454--1459, 2019.

\bibitem{Nayef2017ICDAR2017RR}
Nibal Nayef, Fei Yin, Imen Bizid, Hyunsoo Choi, Yuan Feng, Dimosthenis
  Karatzas, Zhenbo Luo, Umapada Pal, Christophe Rigaud, Joseph Chazalon, Wafa
  Khlif, Muhammad~Muzzamil Luqman, Jean-Christophe Burie, Cheng-Lin Liu, and
  Jean-Marc Ogier.
\newblock Icdar2017 robust reading challenge on multi-lingual scene text
  detection and script identification - rrc-mlt.
\newblock In {\em 2017 14th IAPR International Conference on Document Analysis
  and Recognition (ICDAR)}, volume~01, pages 1454--1459, 2017.

\bibitem{Petroni2019LanguageMA}
Fabio Petroni, Tim Rockt{\"a}schel, Patrick Lewis, Anton Bakhtin, Yuxiang Wu,
  Alexander~H. Miller, and Sebastian Riedel.
\newblock Language models as knowledge bases?
\newblock In {\em EMNLP}, page 1772–1791, 2019.

\bibitem{Radford2021LearningTV}
Alec Radford, Jong~Wook Kim, Chris Hallacy, Aditya Ramesh, Gabriel Goh,
  Sandhini Agarwal, Girish Sastry, Amanda Askell, Pamela Mishkin, Jack Clark,
  Gretchen Krueger, and Ilya Sutskever.
\newblock Learning transferable visual models from natural language
  supervision.
\newblock In {\em ICML}, pages 1--16, 2021.

\bibitem{Rao2022DenseCLIPLD}
Yongming Rao, Wenliang Zhao, Guangyi Chen, Yansong Tang, Zheng Zhu, Guan Huang,
  Jie Zhou, and Jiwen Lu.
\newblock Denseclip: Language-guided dense prediction with context-aware
  prompting.
\newblock In {\em 2022 IEEE/CVF Conference on Computer Vision and Pattern
  Recognition (CVPR)}, pages 18061--18070, 2022.

\bibitem{Shi2017DetectingOT}
Baoguang Shi, Xiang Bai, and Serge~J. Belongie.
\newblock Detecting oriented text in natural images by linking segments.
\newblock In {\em 2017 IEEE Conference on Computer Vision and Pattern
  Recognition (CVPR)}, pages 3482--3490, 2017.

\bibitem{Singh2021TextOCRTL}
Amanpreet Singh, Guan Pang, Mandy Toh, Jing Huang, Wojciech Galuba, and Tal
  Hassner.
\newblock Textocr: Towards large-scale end-to-end reasoning for
  arbitrary-shaped scene text.
\newblock {\em 2021 IEEE/CVF Conference on Computer Vision and Pattern
  Recognition (CVPR)}, pages 8798--8808, 2021.

\bibitem{Song2022VisionLanguagePF}
Sibo Song, Jianqiang Wan, Zhibo Yang, Jun Tang, Wenqing Cheng, Xiang Bai, and
  Cong Yao.
\newblock Vision-language pre-training for boosting scene text detectors.
\newblock In {\em CVPR}, pages 15681--15691, 2022.

\bibitem{Tang2019SegLinkDD}
Jun Tang, Zhibo Yang, Yongpan Wang, Qi Zheng, Yongchao Xu, and Xiang Bai.
\newblock Seglink++: Detecting dense and arbitrary-shaped scene text by
  instance-aware component grouping.
\newblock {\em Pattern Recognit.}, 96:106954, 2019.

\bibitem{Tang2022FewCB}
Jingqun Tang, Wenqing Zhang, Hong yi Liu, Mingkun Yang, Bo Jiang, Guan-Nan Hu,
  and Xiang Bai.
\newblock Few could be better than all: Feature sampling and grouping for scene
  text detection.
\newblock In {\em CVPR}, pages 4563--4572, 2022.

\bibitem{Tian2016DetectingTI}
Zhi Tian, Weilin Huang, Tong He, Pan He, and Yu Qiao.
\newblock Detecting text in natural image with connectionist text proposal
  network.
\newblock In {\em ECCV}, pages 1--16, 2016.

\bibitem{Tian2019LearningSE}
Zhuotao Tian, Michelle Shu, Pengyuan Lyu, Ruiyu Li, Chao Zhou, Xiaoyong Shen,
  and Jiaya Jia.
\newblock Learning shape-aware embedding for scene text detection.
\newblock In {\em 2019 IEEE/CVF Conference on Computer Vision and Pattern
  Recognition (CVPR)}, pages 4229--4238, 2019.

\bibitem{Vaswani2017AttentionIA}
Ashish Vaswani, Noam~M. Shazeer, Niki Parmar, Jakob Uszkoreit, Llion Jones,
  Aidan~N. Gomez, Lukasz Kaiser, and Illia Polosukhin.
\newblock Attention is all you need.
\newblock In {\em NeurIPS}, pages 1--11, 2017.

\bibitem{Wan2021SelfattentionBT}
Qi Wan, Haoqin Ji, and Linlin Shen.
\newblock Self-attention based text knowledge mining for text detection.
\newblock In {\em 2021 IEEE/CVF Conference on Computer Vision and Pattern
  Recognition (CVPR)}, pages 5979--5988, 2021.

\bibitem{Wang2020TextRayCG}
Fangfang Wang, Yifeng Chen, Fei Wu, and Xi Li.
\newblock Textray: Contour-based geometric modeling for arbitrary-shaped scene
  text detection.
\newblock In {\em Proceedings of the 28th ACM International Conference on
  Multimedia}, page 111–119, 2020.

\bibitem{Wang2019EfficientAA}
Wenhai Wang, Enze Xie, Xiaoge Song, Yuhang Zang, Wenjia Wang, Tong Lu, Gang Yu,
  and Chunhua Shen.
\newblock Efficient and accurate arbitrary-shaped text detection with pixel
  aggregation network.
\newblock In {\em 2019 IEEE/CVF International Conference on Computer Vision
  (ICCV)}, pages 8439--8448, 2019.

\bibitem{Wang2019ArbitrarySS}
Xiaobing Wang, Yingying Jiang, Zhenbo Luo, Cheng-Lin Liu, Hyunsoo Choi, and
  Sungjin Kim.
\newblock Arbitrary shape scene text detection with adaptive text region
  representation.
\newblock In {\em 2019 IEEE/CVF Conference on Computer Vision and Pattern
  Recognition (CVPR)}, pages 6442--6451, 2019.

\bibitem{Wang2020ContourNetTA}
Yuxin Wang, Hongtao Xie, Zhengjun Zha, Mengting Xing, Zilong Fu, and Yongdong
  Zhang.
\newblock Contournet: Taking a further step toward accurate arbitrary-shaped
  scene text detection.
\newblock In {\em 2020 IEEE/CVF Conference on Computer Vision and Pattern
  Recognition (CVPR)}, pages 11750--11759, 2020.

\bibitem{Wu2020SynthetictoRealUD}
Weijia Wu, Ning Lu, Enze Xie, Yuxiang Wang, Wenwen Yu, Cheng Yang, and Hong
  Zhou.
\newblock Synthetic-to-real unsupervised domain adaptation for scene text
  detection in the wild.
\newblock In {\em ACCV}, pages 1--14, 2020.

\bibitem{Xie2019SceneTD}
Enze Xie, Yuhang Zang, Shuai Shao, Gang Yu, Cong Yao, and Guangyao Li.
\newblock Scene text detection with supervised pyramid context network.
\newblock In {\em AAAI}, pages 9038--9045, 2019.

\bibitem{Xing2019ConvolutionalCN}
Linjie Xing, Zhi Tian, Weilin Huang, and Matthew~R. Scott.
\newblock Convolutional character networks.
\newblock In {\em 2019 IEEE/CVF International Conference on Computer Vision
  (ICCV)}, pages 9125--9135, 2019.

\bibitem{Xu2021ASB}
Mengde Xu, Zheng Zhang, Fangyun Wei, Yutong Lin, Yue Cao, Han Hu, and Xiang
  Bai.
\newblock A simple baseline for zero-shot semantic segmentation with
  pre-trained vision-language model.
\newblock In {\em ECCV}, 2021.

\bibitem{Xu2019TextFieldLA}
Yongchao Xu, Yukang Wang, Wei Zhou, Yongpan Wang, Zhibo Yang, and Xiang Bai.
\newblock Textfield: Learning a deep direction field for irregular scene text
  detection.
\newblock {\em IEEE Transactions on Image Processing}, 28:5566--5579, 2019.

\bibitem{Xue2019MSRMS}
Chuhui Xue, Shijian Lu, and Wei Zhang.
\newblock Msr: Multi-scale shape regression for scene text detection.
\newblock In {\em IJCAI}, pages 989--995, 2019.

\bibitem{Xue2022LanguageMA}
Chuhui Xue, Wenqing Zhang, Yu Hao, Shijian Lu, Philip H.~S. Torr, and Song Bai.
\newblock Language matters: A weakly supervised vision-language pre-training
  approach for scene text detection and spotting.
\newblock In {\em ECCV}, pages 1--19, 2022.

\bibitem{Yao2014AUF}
Cong Yao, Xiang Bai, and Wenyu Liu.
\newblock A unified framework for multi-oriented text detection and
  recognition.
\newblock {\em IEEE Transactions on Image Processing}, 23:4737--4749, 2014.

\bibitem{yao2012detecting}
Cong Yao, Xiang Bai, Wenyu Liu, Yi Ma, and Zhuowen Tu.
\newblock Detecting texts of arbitrary orientations in natural images.
\newblock In {\em 2012 IEEE conference on computer vision and pattern
  recognition}, pages 1083--1090. IEEE, 2012.

\bibitem{Ye2020TextFuseNetST}
Jian Ye, Zhe Chen, Juhua Liu, and Bo Du.
\newblock Textfusenet: Scene text detection with richer fused features.
\newblock In {\em International Joint Conference on Artificial Intelligence},
  2020.

\bibitem{Zhan2019GADANGD}
Fangneng Zhan, Chuhui Xue, and Shijian Lu.
\newblock Ga-dan: Geometry-aware domain adaptation network for scene text
  detection and recognition.
\newblock In {\em 2019 IEEE/CVF International Conference on Computer Vision
  (ICCV)}, pages 9104--9114, 2019.

\bibitem{Zhang2019LookMT}
Chengquan Zhang, Borong Liang, Zuming Huang, Mengyi En, Junyu Han, Errui Ding,
  and Xinghao Ding.
\newblock Look more than once: An accurate detector for text of arbitrary
  shapes.
\newblock In {\em 2019 IEEE/CVF Conference on Computer Vision and Pattern
  Recognition (CVPR)}, pages 10544--10553, 2019.

\bibitem{Zhang2020DeepRR}
Shi-Xue Zhang, Xiaobin Zhu, Jie-Bo Hou, Chang Liu, Chun Yang, Hongfa Wang, and
  Xu-Cheng Yin.
\newblock Deep relational reasoning graph network for arbitrary shape text
  detection.
\newblock In {\em 2020 IEEE/CVF Conference on Computer Vision and Pattern
  Recognition (CVPR)}, pages 9696--9705, 2020.

\bibitem{Zhang2016MultiorientedTD}
Zheng Zhang, Chengquan Zhang, Wei Shen, Cong Yao, Wenyu Liu, and Xiang Bai.
\newblock Multi-oriented text detection with fully convolutional networks.
\newblock In {\em 2016 IEEE Conference on Computer Vision and Pattern
  Recognition (CVPR)}, pages 4159--4167, 2016.

\bibitem{Zhou2022ConditionalPL}
Kaiyang Zhou, Jingkang Yang, Chen~Change Loy, and Ziwei Liu.
\newblock Conditional prompt learning for vision-language models.
\newblock In {\em CVPR}, pages 16816--16825, 2022.

\bibitem{Zhou2021LearningTP}
Kaiyang Zhou, Jingkang Yang, Chen~Change Loy, and Ziwei Liu.
\newblock Learning to prompt for vision-language models.
\newblock {\em International Journal of Computer Vision}, page 2337–2348,
  2022.

\bibitem{Zhou2017EASTAE}
Xinyu Zhou, Cong Yao, He Wen, Yuzhi Wang, Shuchang Zhou, Weiran He, and Jiajun
  Liang.
\newblock East: An efficient and accurate scene text detector.
\newblock In {\em 2017 IEEE Conference on Computer Vision and Pattern
  Recognition (CVPR)}, pages 2642--2651, 2017.

\bibitem{Zhu2017UnpairedIT}
Jun-Yan Zhu, Taesung Park, Phillip Isola, and Alexei~A. Efros.
\newblock Unpaired image-to-image translation using cycle-consistent
  adversarial networks.
\newblock In {\em 2017 IEEE International Conference on Computer Vision
  (ICCV)}, pages 2242--2251, 2017.

\bibitem{zhu2021fourier}
Yiqin Zhu, Jianyong Chen, Lingyu Liang, Zhuanghui Kuang, Lianwen Jin, and Wayne
  Zhang.
\newblock Fourier contour embedding for arbitrary-shaped text detection.
\newblock In {\em 2021 IEEE/CVF Conference on Computer Vision and Pattern
  Recognition (CVPR)}, pages 3122--3130, 2021.

\end{thebibliography}
}

\clearpage

\newpage
\appendix

\section{Appendix}

\subsection{Datasets}
\noindent\textbf{ICDAR2013}~\cite{Karatzas2013ICDAR2R} is high-resolution English dataset for focused scene text detection, including 229 images for training and 233 images for testing.

\noindent\textbf{ICDAR2015}~\cite{karatzas2015icdar} is a multi-oriented text detection dataset for English text that includes 1,000 training images and 500 testing images. Scene text images in this dataset were taken by Google Glasses without taking care of positioning, image quality, and viewpoint.

\noindent\textbf{MSRA-TD500}~\cite{yao2012detecting} is a multi-language dataset that includes English and Chinese, including 300 training images and 200 testing images. We also include extra 400 training images from HUST-TR400~\cite{Yao2014AUF} following the previous methods~\cite{Liao2020RealtimeST,Zhou2017EASTAE}.

\noindent\textbf{CTW1500}~\cite{liu2019curved} consists of 1,000 training images and 500 testing images which focuses on the curved text. The text instances are annotated in the text-line level by polygons with 14 vertices.

\noindent\textbf{Total-Text}~\cite{ch2019total} contains 1,255 training images and 300 testing images. The text instances are labeled at the word level. It includes horizontal, multi-oriented, and curved text shapes.

\noindent\textbf{ArT}~\cite{chng2019icdar2019} includes 5,603 training images and 4,563 testing images. It is a large-scale multi-lingual arbitrary-shape scene text detection dataset. The text regions are annotated by the polygons with an adaptive number of key points. Note that it contains Total-Text and CTW1500.

\noindent\textbf{MLT17}~\cite{Nayef2017ICDAR2017RR} includes 9 languages text representing 6 different scripts annotated by quadrangle. It has 7,200 training images, 1,800 validation images, and 9,000 testing images. We use both the training set and the validation set in the finetune period.

\noindent\textbf{MLT19}~\cite{nayef2019icdar2019} is a large-scale multi-lingual scene text detection datasets. It contains 10,000 training images and 10,000 testing images, and labeled at word level.

\noindent\textbf{SynthText}~\cite{gupta2016synthetic} It contains 800k synthetic images generated by blending natural images with artificial text, which are all word-level annotated.

\noindent\textbf{TextOCR}~\cite{Singh2021TextOCRTL} is a large-scale high quality scene text datasets collected from Open Images\footnote{\url{https://storage.googleapis.com/openimages/web/index.html}}. It contains 30 words on average per image. It has 24,902 training images and 3,232 testing images, and is annotated with polygons.

\subsection{More Quantitative Results}

\noindent\textbf{Multi-lingual Real-to-real Adaptation.}
We conducted multi-lingual generalization ability experiments as shown in Table~\ref{tab:real_to_real_mlt17_19}. The results show that the pluggable TCM can also benefit to multi-lingual scenarios text detection via leveraging the pretrained knowledge of CLIP, which demonstrates the effectiveness of our method for domain adaptation.
	
	\begin{table}[ht]
		\centering
	
	\newcommand{\tabincell}[2]{\begin{tabular}{@{}#1@{}}#2\end{tabular}}
\begin{tabular}{lc}
    \toprule
	\multirow{1}[0]{*}{Method} &
	\multicolumn{1}{c}{MLT17 $\rightarrow$ MLT19} 
	 \\
    \midrule
	DBNet~\cite{Liao2020RealtimeST}  & 47.4  \\
    \midrule
	TCM-DBNet & \textbf{67.5}    \\
    \bottomrule
\end{tabular}
		\caption{Real-to-real adaptation. F-measure (\%) is reported.
		}
		\label{tab:real_to_real_mlt17_19}
	\end{table}

\noindent\textbf{Ablation Study for the Different Predefined Language Prompt.} We conducted ablation study on the predefined language prompt with different string using TCM-DBNet in Table~\ref{tab:differ_predefined_prompt}. Results show that without predefined language prompt the performance is harmed. In addition, it can be seen that there is little performance variation with different predefined language prompt. 

\begin{table}[htbp]
\centering
\normalsize
\setlength\tabcolsep{2.3pt}
{
\begin{tabularx}{1.0\linewidth}{lc}
\hline
Predefined language prompt       & IC15   \\
\hline
``Text''         &   89.2        \\
``A set of arbitrary-shape text instances'' & 89.0 \\ 
``The pixels of many arbitrary-shape text instances'' & 88.9 \\
without predefined language prompt & 87.7 \\
\hline
\end{tabularx}}
\vspace{-0.5em}
\caption{Ablation study of the different predefined language prompt.}
\label{tab:differ_predefined_prompt}
\end{table}

\noindent \textbf{Ablation Study for Training with Large-scale Dataset.} We conducted ablation study of training TCM-DBNet on IC15 with extra TextOCR~\cite{Singh2021TextOCRTL} data. As shown in Table~\ref{tab:train_textocr}, when using additional large-scale TextOCR as training data, our model can achieve further improvement, suggesting the compatibility of our method with large-scale datasets.

\begin{table}[htbp]
\centering
\normalsize
{
\begin{tabularx}{0.75\linewidth}{llc}
\hline
Model        & Training data & F (\%)  \\
\hline
TCM-DBNet        &   IC15        & 89.2          \\
TCM-DBNet   &    IC15+TextOCR      &  90.4       \\
\hline
\end{tabularx}}
\vspace{-0.5em}
\caption{Ablation study of training TCM-DBNet on IC15 with extra TextOCR data.}
\label{tab:train_textocr}
\end{table}

\noindent \textbf{Ablation study for CLIP Backbone Generalization.} We conducted ablation study to investigate the generalization performance of DBNet by directly replacing the backbone of DBNet with CLIP backbone, as shown in Table~\ref{tab:clip_r50}. It shows that the CLIP-R50 can indeed bring benefit for generalization. However, integrating with TCM, the performance can be significantly improved. It suggests that directly using the pre-trained CLIP-R50 is not strong enough to improve the generalization performance of the existing text detector, which further indicates that synergistic interaction between the detector and the CLIP is important.

\begin{table}[htbp]
\centering
\normalsize
{
\begin{tabularx}{1.0\linewidth}{lccc}
\hline
Model        & Backbone & ST $\rightarrow$ IC13 &  ST $\rightarrow$ IC13 \\
\hline
DBNet        &   R50       & 71.7   & 64.0        \\
DBNet   &    CLIP-R50      &  73.1    &   67.4     \\
TCM-DBNet   &    CLIP-R50      &  79.6    & 76.7       \\
\hline
\end{tabularx}}
\vspace{-0.5em}
\caption{Ablation study on CLIP backbone. R50 means ResNet50.}
\label{tab:clip_r50}
\end{table}

\subsection{More Visualization Results}
\begin{figure}[htbp]
\centering
\includegraphics[width=0.2\textwidth]{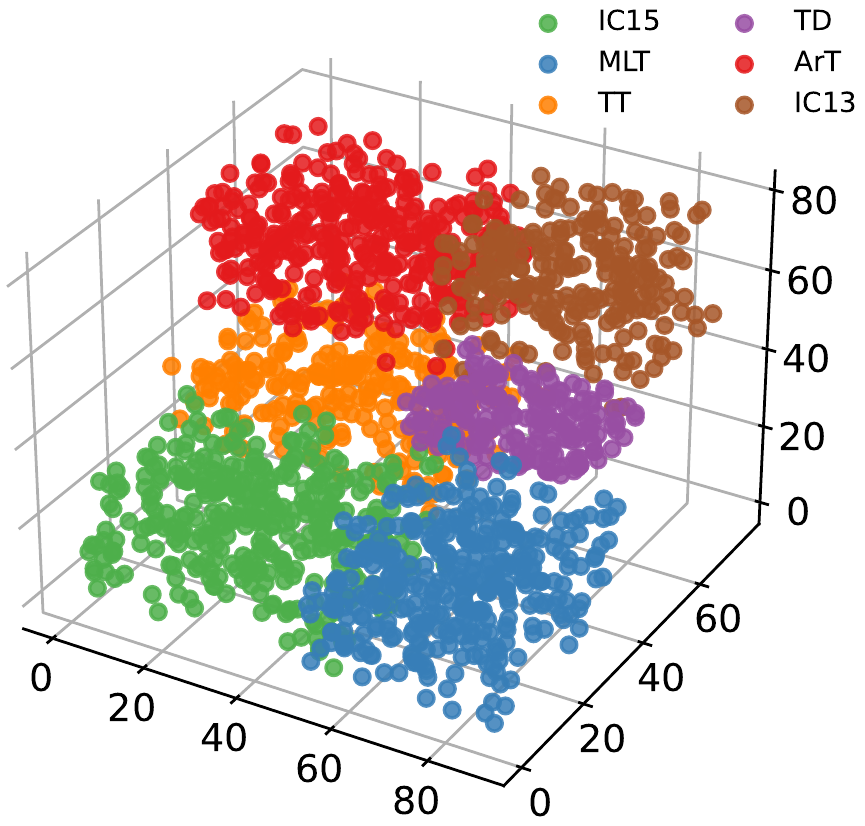}
\caption{t-SNE of conditional cue ($\mathbf{cc}$). MLT short for MLT17.}
\vspace{-1em}
\label{fig:conditiona_cue}
\end{figure}

\noindent\textbf{Conditional Cue.} We visualize the t-SNE of the generated conditional cue ($\mathbf{cc}$) on six datasets, as illustrated in Fig.~\ref{fig:conditiona_cue}. The structured distribution indicates our model has learned the distribution of every domain dataset in high-dimensional feature space, which is useful for improving the generalization ability. 

\begin{figure*}[t]
\centering
\includegraphics[width=0.99\textwidth]{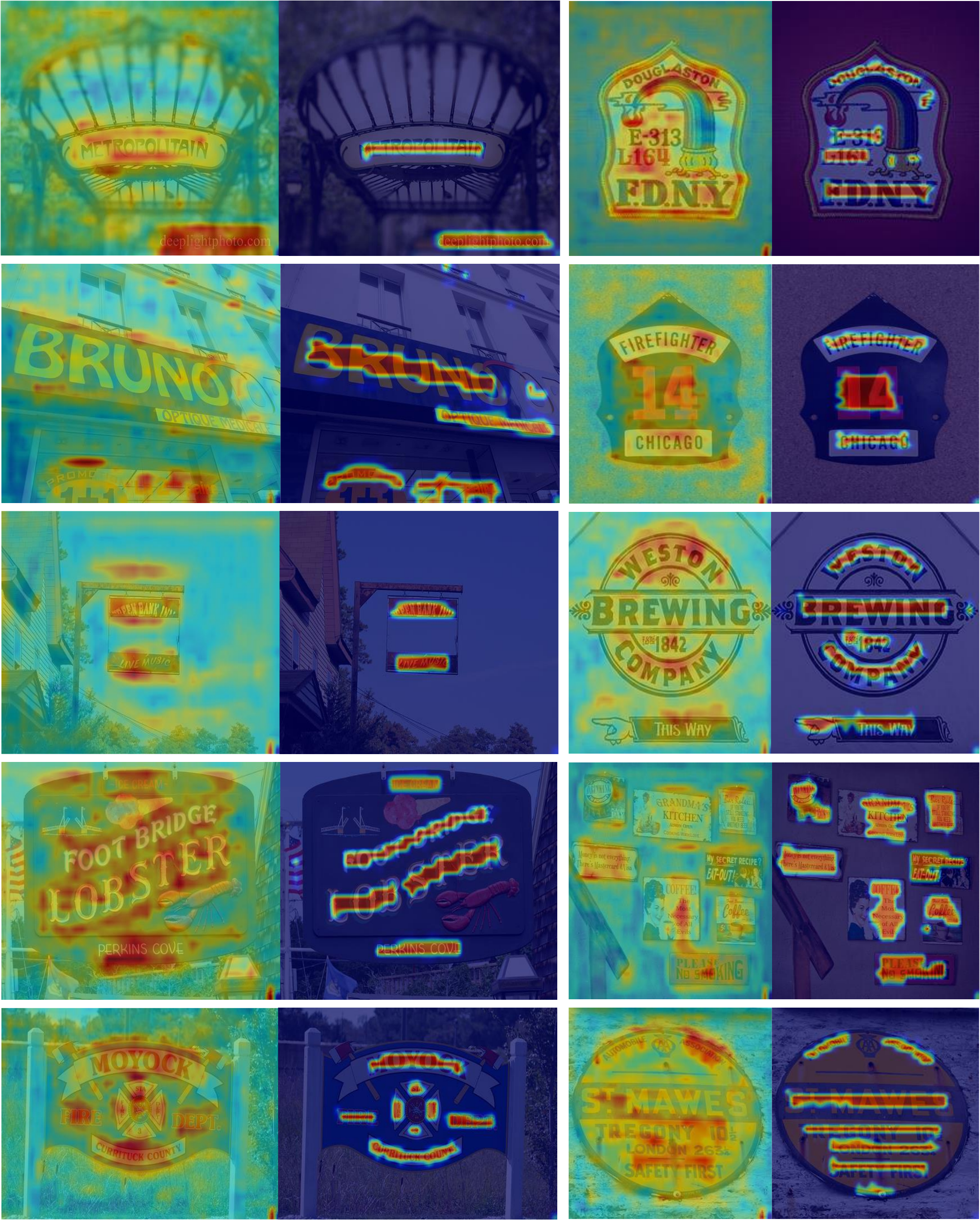}
\caption{Visualization results of our method on CTW1500. For each pair, the left is the image embedding $\bm{I}$, and the right is the generated visual prompt $\tilde{\bm{I}}$.  Best view in screen.}
\label{fig:appendix_ctw}
\end{figure*}

\begin{figure*}[t]
\centering
\includegraphics[width=0.99\textwidth]{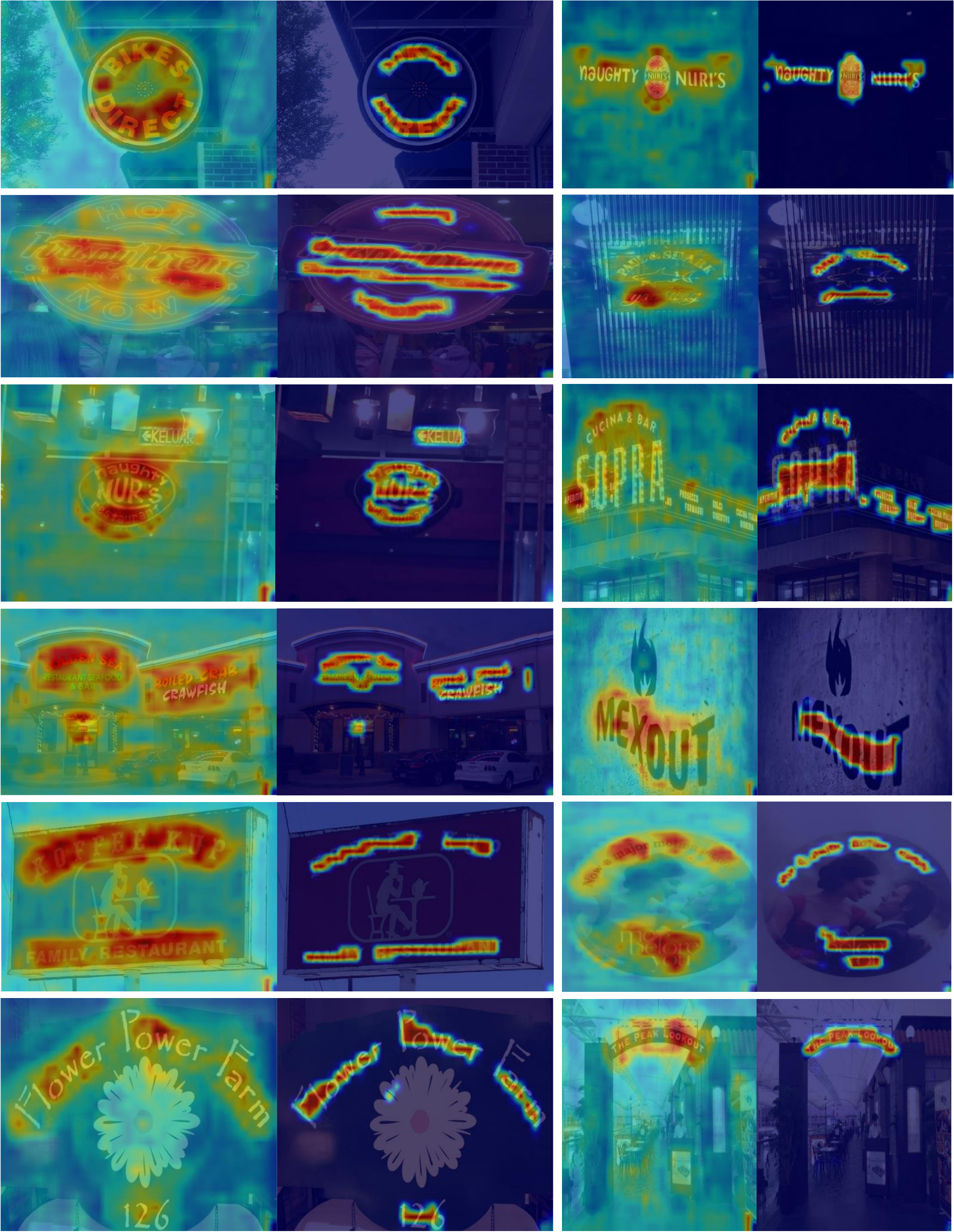}
\caption{Visualization results of our method on Total-Text. For each pair, the left is the image embedding $\bm{I}$, and the right is the generated visual prompt $\tilde{\bm{I}}$.  Best view in screen.}
\label{fig:appendix_tt}
\end{figure*}

\begin{figure*}[t]
\centering
\includegraphics[width=0.99\textwidth]{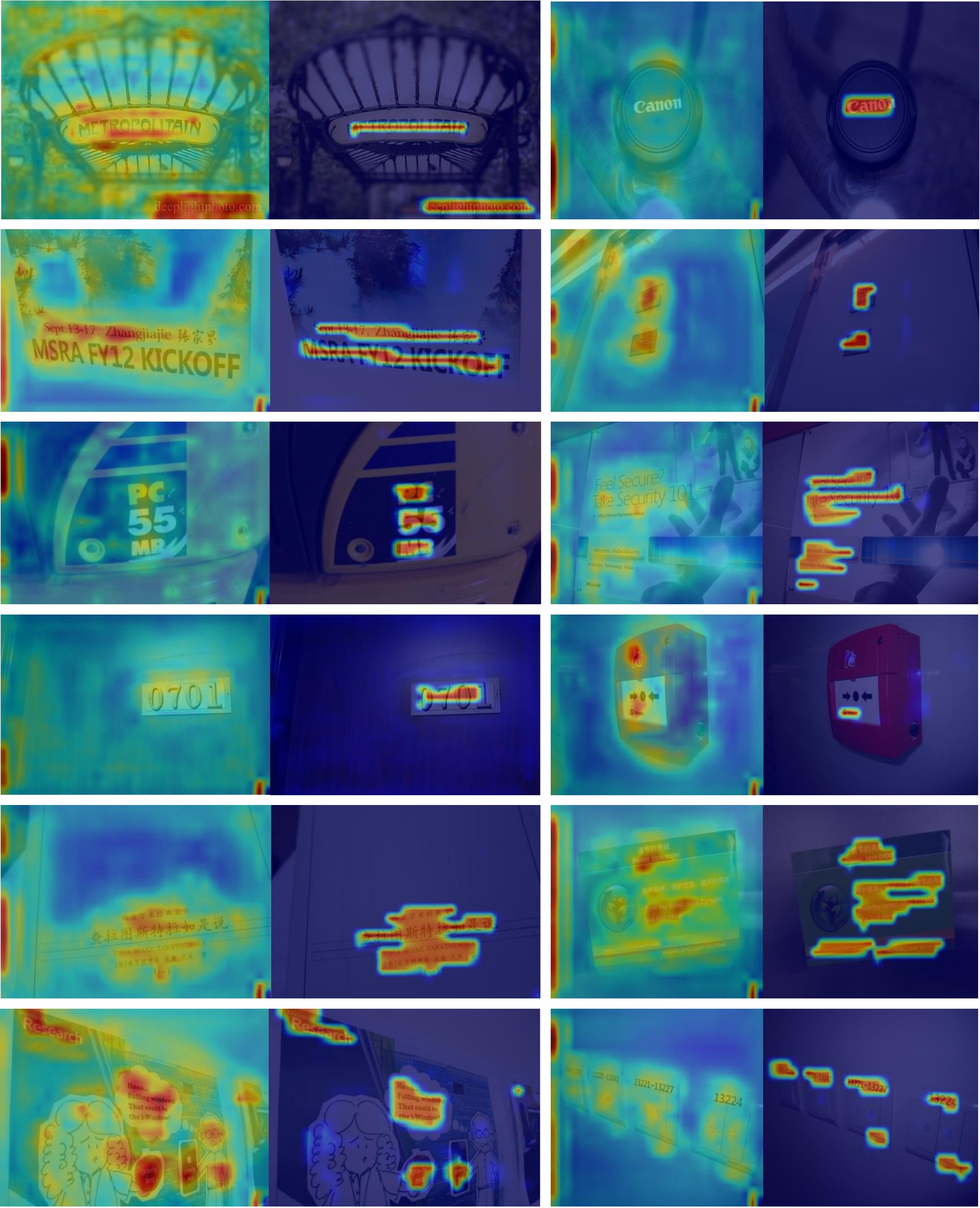}
\caption{Visualization results of our method on MSAR-TD500. For each pair, the left is the image embedding $\bm{I}$, and the right is the generated visual prompt $\tilde{\bm{I}}$.  Best view in screen.}
\label{fig:appendix_td}
\end{figure*}

\begin{figure*}[t]
\centering
\includegraphics[width=0.99\textwidth]{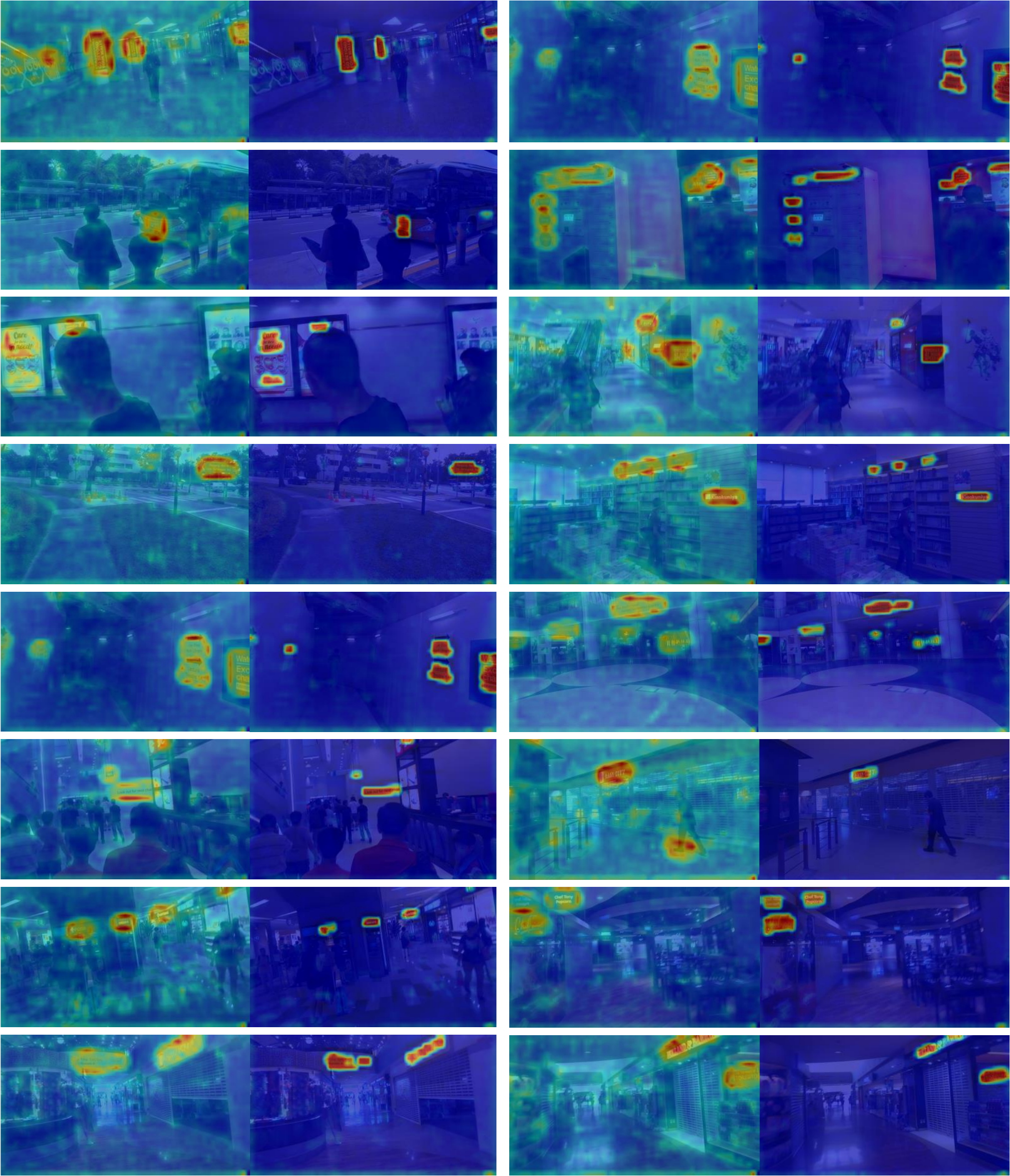}
\caption{Visualization results of our method on ICDAR2015. For each pair, the left is the image embedding $\bm{I}$, and the right is the generated visual prompt $\tilde{\bm{I}}$.  Best view in screen.}
\label{fig:appendix_ic15}
\end{figure*}

\noindent\textbf{Visual Prompt.} Fig.~\ref{fig:appendix_ctw} - Fig.~\ref{fig:appendix_ic15} are more qualitative results of the image embedding $\bm{I}$ and the generated visual prompt $\tilde{\bm{I}}$ on CTW1500, Total-Text, MSRA-TD500, and ICDAR2015, respectively. The visual prompt $\tilde{\bm{I}}$ has contains fine-grained information of text regions.

\end{document}